 \documentclass[5p,times]{elsarticle}
\usepackage{nccmath}
\usepackage{lineno,hyperref}
\modulolinenumbers[5]
\usepackage[ruled,vlined]{algorithm2e}
\usepackage{setspace}
\usepackage{amssymb}
\usepackage{multirow}
\usepackage{multicol}
\journal{Journal of \LaTeX\ Templates}

\begin{document}

\begin{frontmatter}

%\title{Elsevier \LaTeX\ template\tnoteref{mytitlenote}}

\title{Graph Decipher: A transparent dual-attention graph neural network to understand the message-passing mechanism for the node classification}
% \title{Graph Decipher: concern and investigate the optimal graph information efficiently by category in node classification}
% \title{Graph Decipher: understand the message-passing mechanism on graph via graph feature pooling }
% \title{Rethinking the node relevance of the message-passing mechanism on graph in node classification}





\author[mymainaddress]{Yan Pang}
\author[mymainaddress]{Chao Liu\corref{mycorrespondingauthor}}

\cortext[mycorrespondingauthor]{Corresponding author}
\ead{chao.liu@ucdenver.edu}

\address[mymainaddress]{Department of Electrical Engineering, University of Colorado Denver, CO}

\begin{abstract}
Graph neural networks can be effectively applied to find solutions for many real-world problems across widely diverse fields. The success of graph neural networks is linked to the message-passing mechanism on the graph, however the message-aggregating behavior is still not entirely clear in most algorithms. To improve functionality, we propose a new transparent network called Graph Decipher to investigate the message-passing mechanism by prioritizing in two main components: the graph structure and node attributes, at the graph, feature, and global levels on a graph under the node classification task. However the computation burden now becomes the most significant issue because the relevance of both graph structure and node attributes are computed on a graph. In order to solve this issue, only relevant representative node attributes are extracted by graph feature filters, allowing calculations to be performed in a category-oriented manner. Experiments on seven datasets show that Graph Decipher achieves state-of-the-art performance while imposing a substantially lower computation burden under the node classification task. Additionally, since our algorithm has the ability to explore the representative node attributes by category, it is utilized to alleviate the imbalanced node classification problem on multi-class graph datasets.

%In order to reduce the computation burden, the relevance of only representative node attributes, extracted by a graph feature filters, is calculated in a category-oriented way. Graph Decipher can achieve state-of-the-art performance with lower computation under the node classification task, as demonstrated on seven datasets. Besides, an innovative graph data augmentation approach based on our network is utilized to alleviate the imbalanced node classification problem on the multi-class graph datasets.
 
\end{abstract}

\begin{keyword}
%\texttt{elsarticle.cls}\sep \LaTeX\sep Elsevier \sep template
%\MSC[2010] 00-01\sep  99-00
Graph neural network, Message-passing mechanism, Category-oriented, Data Augmentation\end{keyword}

\end{frontmatter}

\nolinenumbers

\section{Introduction}
Graph neural networks (GNNs) offer effective graph-based techniques applied to solve abundant real-world problems in diverse fields, such as social science \cite{hamilton2017inductive}, physical systems \cite{sanchez2018graph, battaglia2016interaction}, protein-protein interaction networks \cite{fout2017protein}, brain neuroscience \cite{goering2020fostering}, knowledge graphs \cite{li2020explain}, etc. The power of current GNNs \cite{hamilton2017inductive, zhou2018graph, tong2020directed, ruiz2020gated, velivckovic2017graph} is largely due to their message-passing mechanisms. However, the underlying behavior that spontaneously aggregates messages on the graph structure is obscure. 

In order to solve this issue, it is crucial to understand the strategy of the message-passing mechanism. This mechanism recursively aggregates information along edges, then updates these newly incorporated features on the center node. Two primary components are involved in this procedure: graph structure and node attributes. Both components need to be clearly identified to view the message-passing mechanism on a graph in node classification tasks. 

In some research \cite{hamilton2017inductive, zhou2018graph, tong2020directed}, messages were passed along edges uniformly without accounting for priority of either graph structure or node attributes. Intuitively, each neighbor node's impact was distinctive to the center node in the node classification task. Thus, attention-based GNNs \cite{ruiz2020gated, velivckovic2017graph} were proposed to further evaluate how the contribution of neighbors to the central node varies according to the graph characteristics. However, the contribution of node attributes was still not identified clearly in these research. Since the node attributes are updated by aggregating the received features from neighbor nodes, the impact of the node attributes is also crucial to affect the transmission of information. Therefore, we hope to find a clear understanding of the roles of both graph structure and node attributes. And particularly in this work, a transparent GNN, Graph Decipher (GD), is proposed to account for the impact of these two main components in the message-passing mechanism on a graph in the node classification tasks.

\begin{figure*}[t]
\begin{center}
\includegraphics[width=15cm]{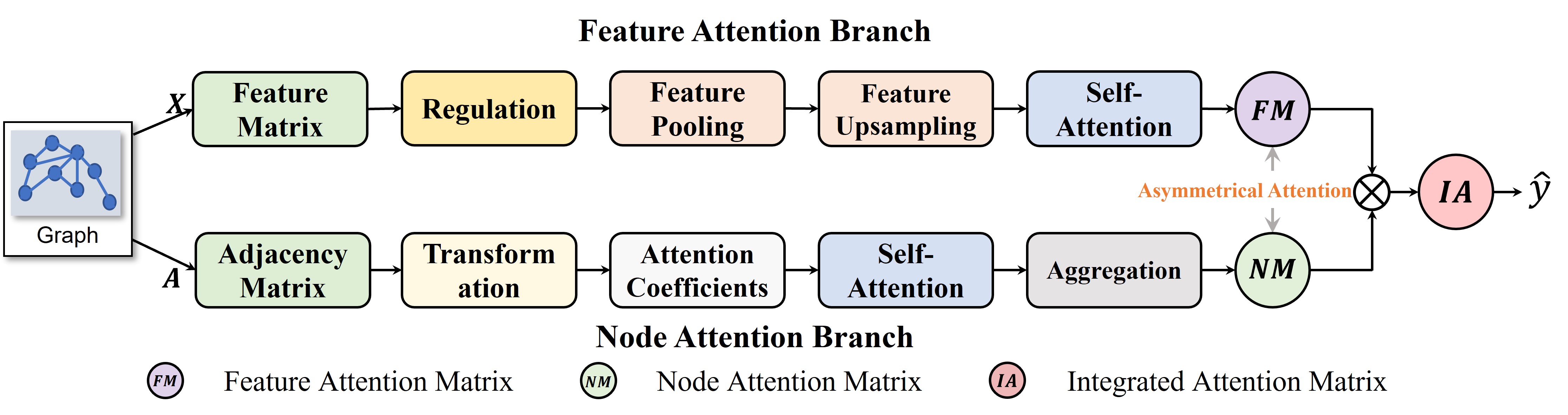}
\end{center}
\caption{The overall architecture of a single-head Graph Decipher includes two main attention branches: Feature Attention Branch (FAB) and Node Attention Branch (NAB). The former branch is dedicated to deep exploiting the importance of node attributes amongst distinct categories of nodes at the feature level, while the latter branch determines the priority of nodes at the graph level.}
\label{fig:top}
\end{figure*}

The proposed GD scrutinizes the message-passing mechanism on the graph from three different perspectives: graph-level, feature-level, and global-level. The graph-level focuses on the distinction of the graph structure in the message-passing mechanism, meanwhile, the feature-level clarifies the contribution of node attributes under the node classification task. As shown in Figure \ref{fig:top}, a single-head of GD contains two parallel branches: a node attention branch (NAB) for graph structure at graph-level and a feature attention branch (FAB) for node attributes at the feature-level. And at the global level, a multi-head attention scheme is used to repeat the computations multiple times in parallel, and then combined together to produce a final decision.
%展示出你这个模型的创新点，也就是别人没有的，而你有的步骤，以及为什么你要这么做。不用特别细节，但需要说的足够清楚。抓住别人眼球告诉读者你在这篇文章中到底要展现什么。
%先说现状怎么是什么，这样造成了什么问题。然后说我们希望得到的效果。因此，我们的解决方案是什么。

% In the original graph, the adjacency matrix is usually used to represent the relationship between nodes in the graph. Still, it ignores the fact that the contributions of neighbors to the central node may be different. In the message-passing mechanism of the graph, we hope that messages from important neighbors can get more attention when they are converged to the central node. Therefore, in this work, a node attention branch is used to calculate the contribution of neighbors to the central node according to the characteristics of the graph task, thereby assigning different weights to the message-passing flows

%The NAB strengthens the contribution of neighbor nodes to the central node at the graph level. Meanwhile, the FAB exploits the attention granted to the node attributes of each node by category at the feature level. 
%总起第一句，你需要让人知道你说的是哪个分支。

The relevance of node attributes or features is first considered at the FAB to gain deeper insights into the message-passing mechanism. Graph feature pooling and upsampling modules are introduced to update the node feature matrix according to the node category. In order to estimate the impact of each node's internal characteristics under the node classification task, a dimension-based self-attention mechanism is proposed, which exploits the attention granted to the node attributes in graph learning. This innovative procedure yields significant improvements in finding the respective optimal attributes of each node according to the categories. Moreover, since only focusing on the optimal attributes instead of all, it also helps reduce the computational burden. Then at the end of the FAB, its output is combined with the concurrent NAB branch that strengthens the contribution of neighbor nodes to the central node for further calculations. Ultimately, a multi-head attention scheme that consists of multiple parallel single-heads outputs the final decisions. Experiments show that this proposed mechanism significantly outperforms the other state-of-the-art work on seven common graph datasets used for node classification tasks: Cora \cite{sen2008collective}, Citeseer \cite{sen2008collective}, PubMed \cite{namata2012query},
Amazon Computers \cite{mcauley2015image}, Amazon Photo \cite{mcauley2015image},
Coauthor CS \cite{shchur2018pitfalls}, and Coauthor Physics \cite{shchur2018pitfalls}, which are illustrated in Table \ref{tab:dataset}.
%就是如何降低计算量/读者读到这的时候一定会提出质疑。最好在上一段末就直接指出计算量大的问题，并简述解决方案。这样才不会让别人不愿意读下去。

Since GD not only focuses on the graph structure but also considers the contribution of node attributes to message passing, it can also perform well on the imbalanced node classification tasks. For datasets with inhomogeneous distribution of node categories, the network prefers the majority classes with a large amount of data, while samples from a minority of categories are under-represented, resulting in sub-optimal performance. A possible solution is to balance entries in each category of the dataset by increasing the number of nodes in the minorities. To avoid the over-fitting problem, only the dominant node attributes of each category analyzed by GD are retained, and the unrepresentative features are randomly dropped or replaced during the reproduction process. Experiments show that this proposed data augmentation method significantly improves the performance on the imbalanced node classification tasks.

% 要说出来到底用的怎样的算法，方法去做了data augmentation，不要藏着掖着。别人光做这一件事情就能发个文章了。你这边要简短但非常明确的说出来，怎么去做的。也就是别人做不到，而你的算法能办到的关键核心是啥
% With the transparent message-passing mechanism, the relatively essential node attributes of the minorities of the imbalanced graph datasets can be pre-analyzed according to the category under the node classification task. These optimal-related node attributes of minorities benefit the network performance by balancing the distribution of node categories.

%utilized to reproduce to balance the multi-class dataset, which improves the accuracy of the network to a certain extent. 
% Contributions不能并列。毕竟你这文章的主题是提出一个新模型GD。那contribution应该是这个模型本身。至于在应用上的良好效果，自然要提，但主次要分明。
The contributions in this paper are summarized as follows:
\begin{itemize}
\item In order to improve the performance of the GNN model on node classification tasks, a new perspective, that is, to improve the clarity of the message passing mechanism on the graph is recommended.
\item A transparent GNN, Graph Decipher, is proposed. This scheme can explain how the graph structure and node attributes affect the message passing mechanism in the node classification tasks from the three levels of feature, graph, and global.
\item Unlike the common methods that assign the same weight to each feature of the same node, we designed novel graph feature pooling and upsampling modules to extract and pay more attention to the dominant features for optimizing the message passing mechanism.
\item To reduce the computational burden, we analyze node attributes in groups by category, and only the representative attributes extracted by the graph feature pooling filter are utilized in the calculation.
\item Since GD has the ability to perform representative analysis on the features of each node, it can be used to augment the samples from a minority of categories, thereby improving the performance on the imbalanced node classification tasks.
\end{itemize}

The structure of this paper is organized as follows: Section \ref{RW} presents a review of the latest work on graph neural networks and multi-class imbalanced graph learning. Section \ref{PR} identifies key terminology and basic concepts used in this paper. Section \ref{ME} describes the detailed mathematical process of GD from three different perspectives. The experiment design and results are discussed in Section \ref{EX}. Finally, Section \ref{CON} presents our conclusion.

\begin{table*}[h]
\centering
\footnotesize
\caption{General Notations}
\vspace{0.5em}
\begin{tabular}{c|c|c|c}
\hline\noalign{\footnotesize}
Notations & Description & Notations & Description \\
\noalign{\footnotesize}\hline\noalign{\footnotesize}
G & Graph & A & Adjacency Matrix \\
E & Edges / Connections & $e_{ij} $ & Edge between Node i and j\\
N & Nodes / Objects & v & Center Node \\
$\boldsymbol{X}$ & Node Feature Vectors & $F$ & Node Feature Dimension \\
$\boldsymbol{h}$ & Hidden Representation & l & Neural Network Layer \\
$\alpha$ & Feature Attention Coefficient & $\beta$ & Node Attention Coefficient \\
y & Ground Truth Labels  & $\hat{y}$ & Predicted Labels\\
$\theta$ & Parameters of Graph Model & $\theta^{*}$ & Optimal Parameters of Graph Model \\
$f_{\theta}$ & Prediction of Graph Model & L & Loss Function  \\ 
l($\cdot$,$\cdot$) & Pair-wise loss function & $\parallel \cdot \parallel$ & $l_{0}$ norm \\
\noalign{\footnotesize}\hline
\end{tabular}
\label{tab:notations}
\end{table*}

 \section{Related Work}\label{RW}
 \subsection{Graph Neural Networks}
The general understanding of the message-passing mechanism of GNNs is that it updates the feature representation of all nodes by aggregating messages along edges on a graph where the contribution of graph structure and node attributes is crucial. 

Recent researches \cite{wu2020comprehensive, hammond2011wavelets} apply aggregation operations directly to graphs and aggregate messages with shared weights from neighbors to each center node. This kind of GNNs only considers passing messages uniformly from one- or two-hop neighbors along edges. GraphSAGE \cite{hamilton2017inductive} aggregates and updates features in a range of the two-hop neighbors to the center node. DGCN \cite{tong2020directed} considers the first- and second-order proximity to aggregate the attributes on the directed graphs. However, these message aggregators collect information from all neighbors equally, ignoring the relative importance of different neighbor nodes. 

In order to solve the issue as mentioned above, more studies \cite{velivckovic2017graph, zhang2018gaan, cirstea2021graph} consider the attention-based architecture to compute the hidden representations of each node in the graph. By calculating the node attention coefficients from the neighbors before reaching the center node, GAT \cite{velivckovic2017graph} implicitly specifies the relevance of neighbor nodes in the node classification task. GaAN \cite{zhang2018gaan} considers the priority of the multi-head attention with a convolution sub-network. WGCN \cite{zhao2021wgcn} utilizes weighted structural features to explore directional structural information for nodes. However, these GNNs only concentrate at the graph-level by assigning arbitrary weights to the neighbor nodes. Although the neighbor nodes' priority is investigated, the relevance of node features is still ignored. 

The dimensions of node attributes are supposed to play different roles in the above GNNs under the node classification task. For example, at nodes marked with lower weights, significant internal attributes will be suppressed to propagate. Conversely, nodes with higher attention coefficients may have inconsequential attributes amplified and passed along to the central node, which causes information interference that limits the overall accuracy of the network in the node classification task. 

Our proposed GD exploits deep characteristics of the message-passing mechanism on both graph structure and node attributes. The transparent mechanism allows straightforward investigation of each node's internal and external impacts in a graph under the node classification task. These transparent nodes can also be utilized in a wide range of applications, such as social networks, recommended systems, the internet of things, emotion estimation, etc.

%Filtering neighbors with the least priority in a substantial social network graph contributes to graph network compression efficiency.

\subsection{Graph Data Augmentation}\label{rmi}

Node classification is a primary graph task for a wide range of applications \cite{hamilton2017inductive, sanchez2018graph, battaglia2016interaction, fout2017protein, goering2020fostering, li2020explain}, as it determines the category of the central node by comparing neighbors based on the message-passing mechanism which learns the node attributes in the multi-class graph dataset. The distribution of multi categories in the graph dataset is directly tied to the network's overall performance. Many researches \cite{drummond2003c4, chawla2002smote, japkowicz2002class, shi2020multi} have demonstrated that neural networks are more inclined to learn features from categories with larger amounts of data, which results in relatively lower accuracy of minor categories. Therefore, it is critical to augmenting the graph data to achieve a balanced distribution of nodes by category.

%先说现状怎么是什么，这样造成了什么问题。然后说我们希望得到的效果。因此，我们的解决方案是什么。
%The adjacency matrix is usually used to represent the relationship between nodes in the graph but ignores the fact that the contributions of neighbors to the central node may be different. In the message-passing mechanism of the graph, we hope that messages from important neighbors can get more attention when they are converged to the central node. Therefore, in this work, a NAM module is used to calculate the contribution of neighbors to the central node according to the characteristics of the graph task, thereby assigning different weights to the message-passing flows.

Rong et al. \cite{rong2019dropedge} designed a DropEdge technology to randomly deletes some edges before each training epoch to prevent messages from being passed from the nodes labeled as the majority category. Chen \cite{chen2020measuring} proposed to change the connections between nodes by adding edges to nodes of the same category or disconnecting nodes from different categories. Although the data-imbalance problem can be alleviated, this approach may lead to propagation errors on the modified graph. Shi \cite{shi2020multi} facilitated the partition of the annotated nodes with a class-conditioned adversarial strategy in the training process. However, these approaches did not actually increase the number of nodes of the minorities, and their inconsequential attributes may increase the difficulty of model training.

In this work, we employ GD to augment the samples from a minority of categories by performing representative analysis on node attributes to solve the above issues. The dominant and representative node attributes are amplified in the minorities, while the inconsequential ones are suppressed after the data augmentation. Experiments show that such a method significantly improves the performance of the minorities on the imbalanced node classification tasks.

\section{Preliminaries}\label{PR}
This section illustrates the the terminology and preliminary knowledge in this paper. To denote various terminologies, we consider both uppercase and lowercase letters:  bold uppercase letters (e.g., $\bf X$) represent matrices, while lowercase letters (e.g., $x$) stand for vectors and non-bold lowercase letters (e.g., $y$) represent scalars, while Greek letters (e.g.,$\alpha$) indicate parameters as shown in the Table \ref{tab:notations}.

\textsc{Definition 1}. \textbf{General Graph Concept}. \textit{In general, a graph, $G$, contains two main matrices: adjacency matrix, $A$, for graph structure and feature matrix, $\textbf{X} $, for graph information. The element of the adjacency matrix indicates the connection (Edge) of each objects (Nodes). Each row of feature matrix represents the feature representation of one node in the graph. Thus,  $\textbf{X} \subseteq \Re ^{N \times F} $, where $N$ is the number of nodes and $F$ is the node attribute.}

\textsc{Definition 2}. \textbf{Undirected and Directed Graph}. \textit{If the adjacency matrix of one graph is symmetric, this kind of graph is undirected. The edge $e \subseteq E$ is an unordered pair $e = (i, j)$ between node $v_i$ and $v_j$, which means $(v_i,v_j)\equiv (v_j,v_i)$. Unlike undirected graph, the adjacency matrix of directed graph is asymmetric because of edges, $(v_i,v_j)\not\equiv (v_j,v_i)$. The undirected graph is also considered as a directed graph who has bi-directed edges with opposite directions.}

\textsc{Definition 3}. \textbf{Graph Node Classification}. \textit{Given a graph with adjacency matrix $A$ and node feature $\textbf{X} \subseteq \Re ^{N \times d} $, the eventual task is to estimate the anonymous label $y$ of the node $v \subseteq N$ by aggregating and updating the messages from its neighbors. The loss function of the train processing is in Equation \ref{node_classification_loss}.}

\begin{ceqn}
\begin{equation}\label{node_classification_loss}
min L(f_{\theta}(G))=\sum _{i \in V_{L}} l(f_{\theta}(\mathbf{X},A)_{i},y_{i})
\end{equation}
\end{ceqn}

\section{Methods}\label{ME}
% 核心方法需要扩展: 先介绍我们的大致方案是什么，然后具体揭示下两个branch是怎么实现的，接下来再说我们这么做的好处（取得了什么样的成就）
% 这段话其实整个methods中最重要的一段。要写的吸引人的同时明确清晰的总结你的方法，创新点，以及于前人工作的异同
Figure \ref{fig:top} illustrates the main architecture of a single-head of GD. In order to investigate the potential of graphs in the node classification task, two parallel attention branches, node attention branch (NAB) and feature attention branch (FAB), are explored to track down the message-passing mechanism. Like GAT, the NAB learns a node attention matrix, $\boldsymbol{NM}$, which represents the contribution of the neighbors to the central node at the graph-level. While an innovative FAB module is utilized to obtain the feature attention matrix ($\boldsymbol{FM}$) to emphasize each node's attributes at the feature-level. Then, the outputs of NAB and FAB ($\boldsymbol{NM}$ and $\boldsymbol{FM}$) are combined to form an integrated attention matrix ($\boldsymbol{IA}$), which contains both attention information of the graph structure and node attributes to complete a single head prediction. The multi-head mechanism is finally used to stabilize the learning process of the node classification task.

% Since the attention information of both nodes and features are deemed, GD improves the convergence speed in the training process.

% The messages from the neighbor nodes are aggregated to the center node v along edges in the original graph. In our mechanism, the spatial information of nodes are sent to the node attention module (NAM). At the end, the priority of each node is determined in the graph task. The distinctive color of edges indicate the different relevance between the neighbor node and the center node v. The details of the NAM is illustrated in the Figure \ref{fig:nam}.

\begin{figure}[t]
\begin{center}
\includegraphics[width=\linewidth]{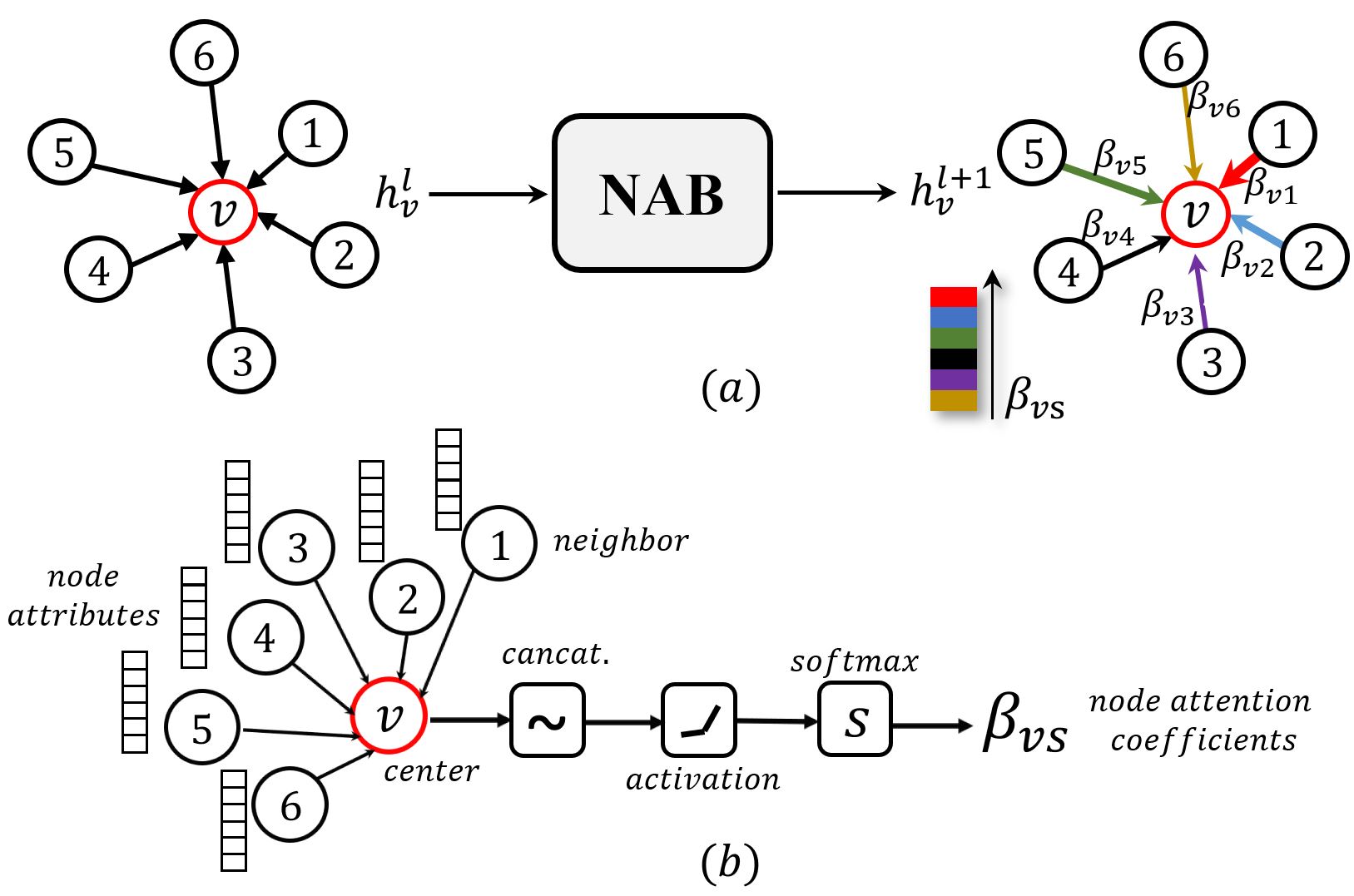}
\end{center}
\caption{Node attention branch. (a) The objective of the NAB is to calculate the relevance of each neighbor to the central node. The different edge colors indicate the priority of these neighbors in the node classification task. (b) The node attention coefficients indicate the significance of the neighbors to the center node in NAB.}
\label{fig:nam}
\end{figure}

\subsection{Graph-Level: Node Attention Branch}\label{nam}
An adjacency matrix is usually used in graph-related tasks to represent the relationship among nodes on the graph. But it ignores the fact that neighbors may contribute differently to the central node. In the message-passing mechanism of the graph, we hope that messages from important neighbors could get more attention when they converge to the central node. Therefore, in this work, a node attention branch is used to calculate the contribution of neighbors to the central node according to the characteristics of the graph task, thereby assigning different weights to the message-passing flows, as shown in Figure \ref{fig:nam} (a).

% 毕竟不是自己的工作。所以介绍的地方有点多。需要明确指出这部分与前人的工作，然后用精炼的语句清楚的介绍这部分。
Figure \ref{fig:nam} (b) illustrates the architecture of the NAB. Similar to GAT, a node self-attention matrix, $\beta_{vs}$, is learned to determine the relevance between neighbors and the center node, $v$, as shown in Equation (\ref{edge}) and (\ref{edge_self_attention}). 

\begin{ceqn}
\begin{equation}\label{edge_self_attention}
\beta_{vs} = softmax(e_{vs})
\end{equation}
\end{ceqn}

\begin{ceqn}
\begin{equation}\label{edge}
e_{vs} = \gamma (W \cdot \mathbf{x_{v}}, W \cdot \mathbf{x_{s}})
\end{equation}
\end{ceqn}

Where $e_{vs}$ represents the importance of neighbor node $s$ to the center node $v$.  $\gamma$ indicates the self-attention mechanism, and $W$ is the weight matrix. $x_{v}$ and $x_{s}$ are the node attributes of the center node $v$ and neighbor node $s$.

%For the center node v, it has a set of neighbor nodes with features $\textbf{X} = \left \{ \textbf{x}^{1}, \textbf{x}^{2},...,\textbf{x}^{n} \right \}$ ,where n is the number of neighbor nodes. In order to obtain sufficient expressive power to transform the neighbor node features into higher-level features, the learnable linear transformation with different attention is adopted. As shown in the Figure \ref{fig:nam}, the self-attention scheme is adopted on the nodes in Equation \ref{edge_self_attention}.

%Where $N_{i}$ is the neighbors of node i, and the edge $e = (i, j)$ between node $v_i$ and $v_j$ is calculated based on the node features and a wight matrix W. As a shared linear transformation, $e = (i, j)$ can be determined in Equation \ref{edge}:

\begin{figure*}[t]
\begin{center}
\includegraphics[width=15cm]{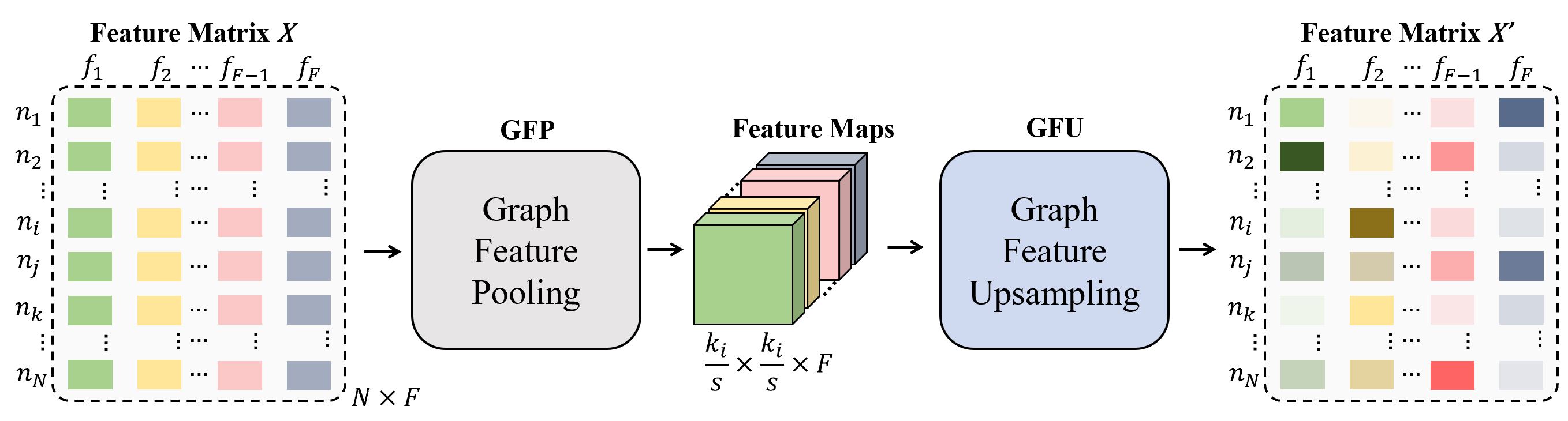}
\end{center}
\caption{The objective of the feature attention branch (FAB) is to obtain the internal priority of node features in a category-oriented way. This branch contains two main modules connected serially: Graph Feature Pooling (GFP) and Graph Feature Upsampling (GFU). The first module picks prominent local attributes on each dimension and reduces the size of the feature maps. The next module calcultes the priority of these highlighted attributes on the updated feature maps. In order to reduce the computation burden, the FAB process is performed by finite node categories. The details of GFP and GFU are illustrated in Figure \ref{fig:fp} and \ref{fig:fu}.}
\label{fig:fum}
\end{figure*}

%The Equation \ref{edge} indicates the significance of message passing from node j to node i. That means the NAM has the ability to make the decision about the magnitude of the consideration from the each neighbor to the center node in the graph task. To achieve the implementation, the NAM is designed as a single-layer feeforward neural network with the parameters $\theta $, while the LeayReLU non-linearity is adopted as the activation function. Thus the Equation \ref{edge_self_attention} can be updated as:

%\begin{ceqn}
%\begin{equation}\label{edge_attention_update}
%\beta_{ij}  =  \frac{exp(LeakyReLU(\theta [W \mathbf{x^{i}} \parallel W %\mathbf{x^{j}}]))}{\sum_{k\subseteq N_{i}exp(LeakyReLU(\theta [W %\mathbf{x^{i}} \parallel W \mathbf{x^{k}}]))}}
%\end{equation}
%\end{ceqn}

%where $\parallel$ indicates the concatenation operation.

Once the attention coefficients are calculated, the activation function $\sigma $ is applied to get the final non-linear node attributes, as shown in Equation (\ref{edge_final}).

\begin{ceqn}
\begin{equation}\label{edge_final}
NAB: \mathbf{x_{v}} = \sigma \left ( \sum_{s\subseteq N_{v}} \beta_{vs} W \cdot \mathbf{x_{s}}\right )
\end{equation}
\end{ceqn}

The node attention coefficients, which represent the contributions of neighbors to the center node in NAB. However, the roles played by the different internal attributes of each node have not been considered. Therefore, a new Feature Attention Branch (FAB) is added to interact with NAB to update the attention of nodes and their corresponding attributes. More details are introduced in section \ref{fam}.

\subsection{Feature-Level: Feature Attention Branch}\label{fam}

The objective of the node classification task is to classify nodes by their attributes. In this task, traditional GNNs first gather the node attributes from the neighbor nodes surrounding the center node. The aggregated attributes are then updated at this targeted node for category prediction. In both procedures, the node attributes are aggregated and updated uniformly on all feature dimensions. However, indiscriminate processing that does not factor in the significance of node attributes may lead to inconsequential node attributes causing redundancy or restricting the desired attributes in the message-passing procedures. Therefore, it is crucial to evaluate the internal priority of node attributes based on the significance of node attributes, allowing representative features to be examined more closely in the message-passing procedure. This strategy improves the efficiency of information transmission and the network's performance under the node classification task.

Ideally, each node's attributes would be thoroughly examined. However, this would cause a steep computation burden. In order to determine the internal priority of attributes and reduce the number of computations, we propose a category-oriented self-attention mechanism. Two modules connected in series constitute the FAB: Graph Feature Pooling (GFP) and Graph Feature Upsampling (GFU), as shown in Figure \ref{fig:fum}. The first module highlights prominent representative attributes and reduces the size of feature maps; the second module distinguishes the priority of these attributes by categories. More details are introduced in section \ref{gfpm} and \ref{gfum}.

\subsubsection{Graph Feature Pooling Module}\label{gfpm}
% 开头要总结一下这个模块是如何起到怎样的作用。细节程度要高于前一段结尾的那个短语
% 在这一段之前，介绍你使用GFP和GFU的另一大原因是减小计算量。用几句话概括的描述减少计算量的核心方法。再在4.2.1的最后对GFP到底如何减少了计算量做一个分析
\begin{figure*}[t]
\begin{center}
\includegraphics[width=18cm]{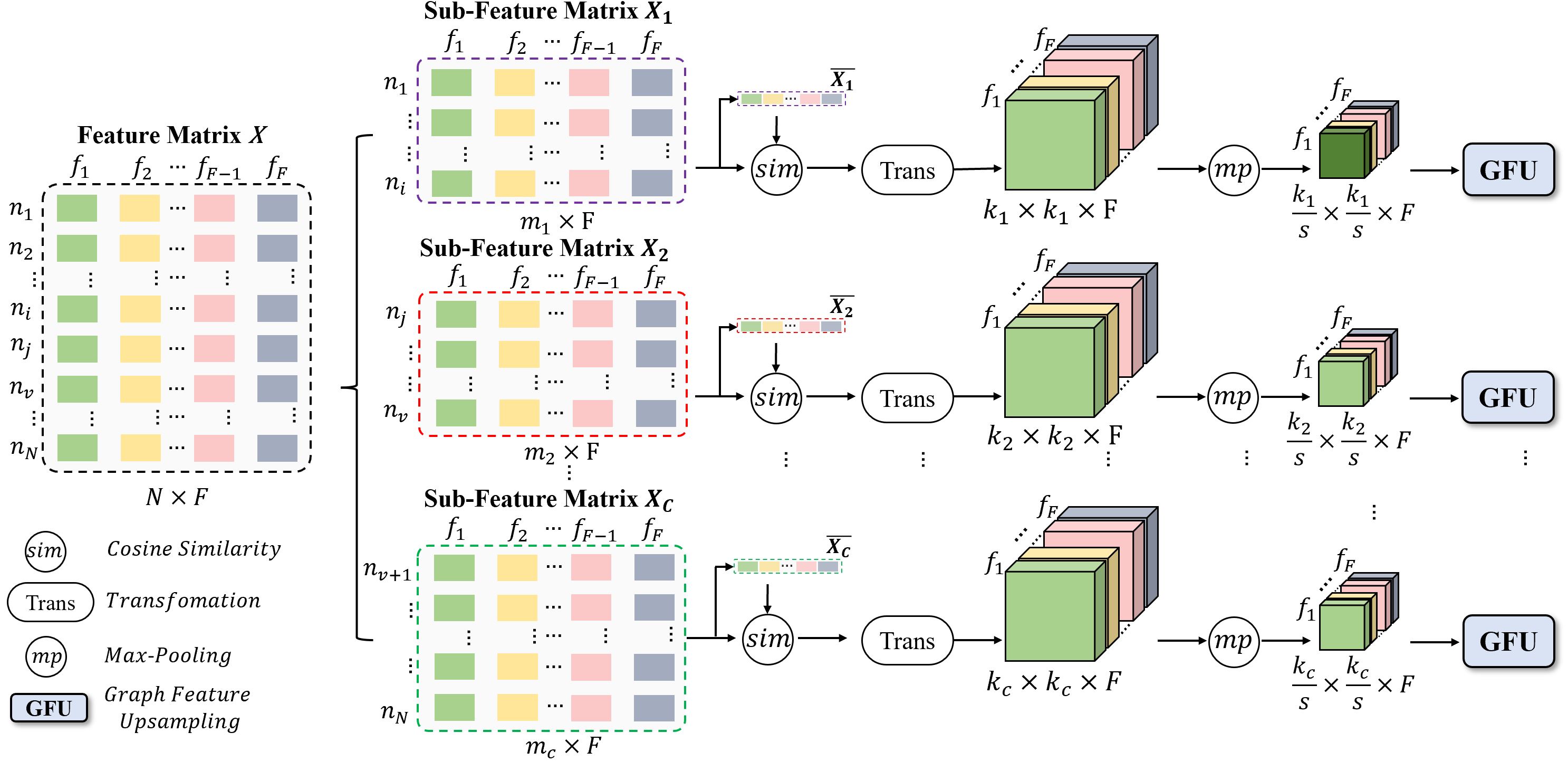}
\end{center}
\caption{The graph feature pooling module (GFP) aims to highlight the local prominent and representative attributes by category. The original 2D feature matrix is split into $C$ sub-feature matrices by node categories. In each sub-matrix, nodes have comparable attributes and are sorted by similarity to the mean feature vector, $\overline{\mathbf{x}}$. Each 2D sub-feature matrix is then transferred to 3D feature maps with the depth of the feature dimension $F$. A max-pooling layer highlights the prominent local attributes on each channel (original feature dimension). These local representative attributes are then sent to the graph feature upsampling (GFU) module to calculate the internal priority of nodes by categories.}
\label{fig:fp}
\end{figure*}

In order to determine the internal priority of attributes in each node, an intuitive method is to calculate the relevance of all node attributes. In this case, the computation burden is equal to $O(\boldsymbol{X}^{n})$, where $\boldsymbol{X} \subseteq \Re ^{n \times F}$ is the feature matrix of the whole graph, and $n$ is the number of the nodes on the graph. This idealized approach can only be applied on the small graph because the computation burden may be much heavier as the number of nodes on a graph increases. However, in practice, we often encounter larger graphs with substantial nodes and edges in areas such as social network graphs.

% 这里少了一句，来引出我们的第一个大创新：By category
We explore the node attributes using category-oriented feature attention coefficients to balance the demands of investigation and categorize computations under the node classification task. From the graph perspective, nodes will be categorized together if they share similar attributes. When the priority of internal attributes of nodes is investigated within each category, the relevance of each attribute can then be sorted under the current task. The most significant attributes can now be selected and sent along graph edges, while the less significant attributes are suppressed in the message-passing mechanism. This priority-based message processing contributes to the performance improvement of GNNs under the node classification task.

Additionally, the number of node categories is finite and far lower than the total number of nodes. Since the focus is given to prioritizing attributes by category instead of individual nodes, the number of feature attention matrices equals the number of node categories on the graph. The values of each feature dimension share the same attention coefficient in each matrix labeled with the same node category. Because the size of the concatenated attention matrices by category is equal to the size of the feature matrix, $\boldsymbol{X}$, of the whole graph, the computation can therefore be reduced to $O(\boldsymbol{X})$ in the learning process.

% 在这一步，我们思考这个问题：图上的众多节点是如何自行分类的？我们首先将节点按照类别进行分类，同时计算了每一个类别下的节点平均feature vector。依据每一个类别下节点对于平均feature vector的相关性，进行排序。
In the GFP, all nodes must first be separated into finite groups. Nodes with the same category are assigned into a single group. According to these newly assigned groups, the original 2D feature matrix $\boldsymbol{X}$ is also divided into $C$ sub-feature matrices as shown in Figure \ref{fig:fp}. The most representative attribute vector from each matrix is the mean feature vector to indicate the corresponding category since nodes in the same group and matrix have comparable attributes. Nodes are sorted in each sub-feature matrix in Equation \ref{similarity} based on the similarity between each node attributes with the mean feature vector. 
%In that case, the new sorted sub-feature matrix boundaries are the lower and upper bounds to indicate the corresponding node category if the quantity of nodes is sufficient.

\begin{ceqn}
\begin{equation}\label{similarity}
sim(\overline{\mathbf{x}}_{i}, \mathbf{x}) = cos\frac{\overline{\mathbf{x}}_{i} \bigodot \mathbf{x}}{\left \|\overline{\mathbf{x}}_{i}  \right \| \cdot  {\left \|\mathbf{x}  \right \|}}
\end{equation}
\end{ceqn}

\begin{ceqn}
\begin{equation}\label{xcave}
\overline{\mathbf{x}}_{i} = \frac{1}{m_{i}}\sum_1^{m_{i}}\mathbf{x}
\end{equation}
\end{ceqn}
Where $\overline{\mathbf{x}}_{i}$ is the mean feature vector, and $m_{i}$ is the number of feature vectors of the $i$-th sub-feature matrix. The symbol $\boldsymbol{x}$ indicates each feature vector of the sub-feature matrix, and the symbol $\bigodot$ represents the dot product between two feature vectors.

Since adjacent nodes represent similar attributes in the sorted sub-feature matrices, the local dominant values must be identified, as they indicate the most representative attributes by dimensions, $F$. In this procedure, each 2D matrix $m_{i} \times F$ is transformed to a corresponding 3D feature map $k_{i} \times k_{i} \times F $ labeled with the node category, where $F$ is the depth of the 3D feature map and also the dimension of 2D feature matrix. Note that the square root of $c_{i}\_n$ is not necessary an integer, thus $k_{i}$ is rounded up to the nearest integer from Equation \ref{3d}. In other words, the shape of the original 2D sub-feature matrix $m_{i}$ is increased to a larger value $k_{i}^{2}$ in most cases. 

%is a pooling operation that calculates the maximum, or largest, value in each patch of each feature map. The results are down sampled or pooled feature maps that highlight the most present feature in the patch,

\begin{ceqn}
\begin{equation}\label{3d}
k_{i}=\lceil\sqrt{m_{i}}\rceil
\end{equation}
\end{ceqn}

\begin{figure*}[t] 
\begin{center} 
\includegraphics[width=18cm]{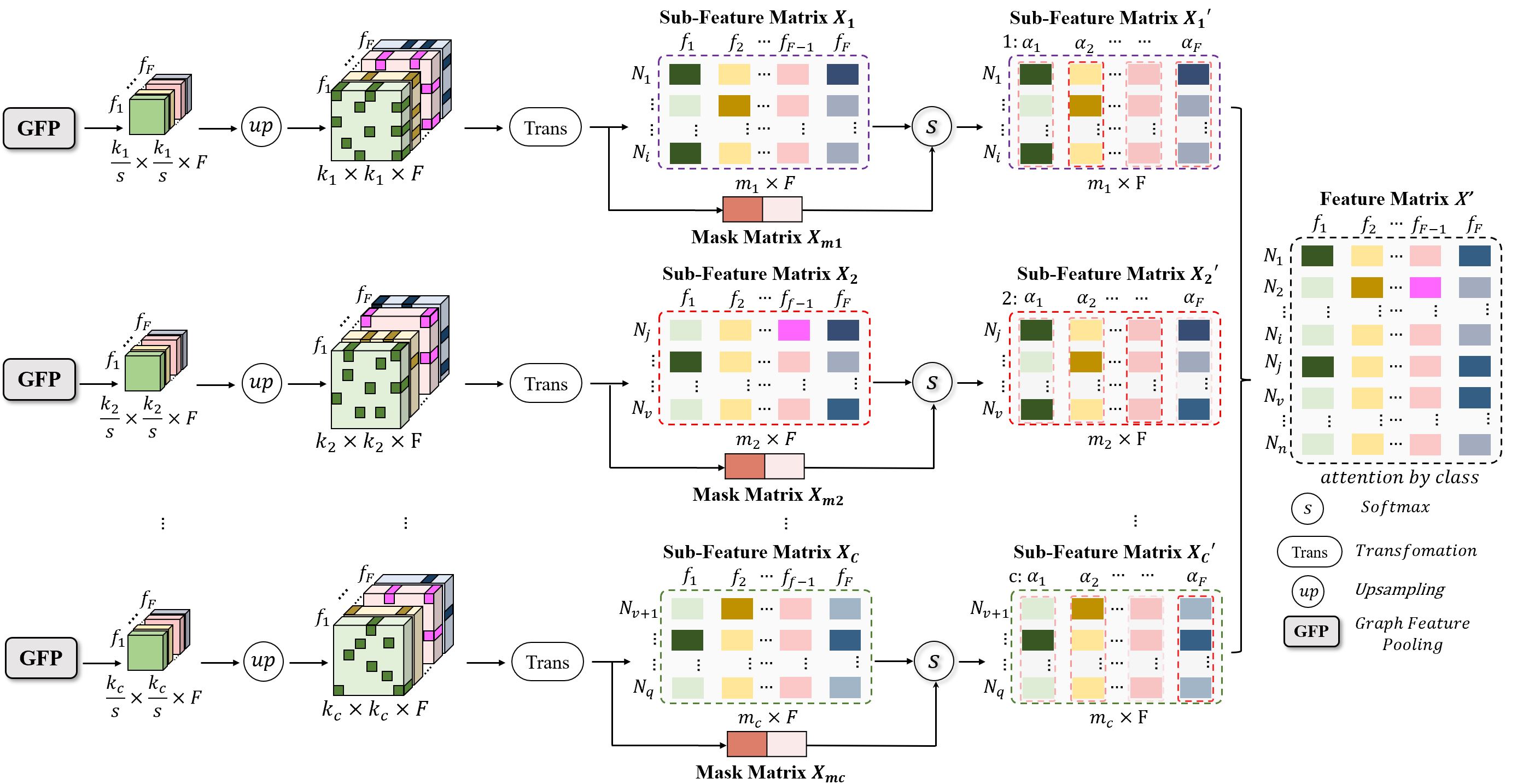} 
\end{center} 
\caption{The objective of the graph feature upsampling (GFU) module is to obtain the internal priority of the local prominent and representative features from the GFP by category. After upsampling and transformation, the 3D feature maps are next converted to 2D sub-feature matrices. A softmax function obtains the internal priority of representative features with the mask with a particular value (0/1) to reduce the computation in each feature matrix. Finally the sub-feature matrices are merged and then converted back to the updated feature matrix $\boldsymbol{X'}$.} 
\label{fig:fu} 
\end{figure*}

The final step of GFP is to highlight the local dominant and representative attributes in the max-pooling operation on each feature map with the corresponding category. The stride of the operation is equal to the size of the pooling filter, and the dimension of the updated feature maps is given by Equation \ref{kfp}.

\begin{ceqn}
\begin{equation}\label{kfp}
dim(feat\_maps) = ( \frac{k_{i}}{s} ,  \frac{k_{i}}{s}, F)
\end{equation}
\end{ceqn}
The stride, $s$, determines the number of representative attributes utilized to calculate the interior priority of nodes in GFU. The computation burden and performance must be balanced, as large stride values lower the computation burden while degrading performance, while lower stride values induce prohibitively enormous computation burdens. This is discussed further in section \ref{fs}.

%上面缺少一句说s的大小可以导致representatives的数量，更大的S可以降低re的总数，这样可以导致后续计算量的进一步降低，但也会造成整体精度的损失。这部分会在5.1讨论。

\subsubsection{Graph Feature Upsampling Module}\label{gfum}
Although the local dominant representative node attributes in each category are highlighted at the end of the GFP, the internal priority of the representative features is not yet known. Thus, the objective of the serially connected module, GFU, is to obtain the internal priority of the locally prominent and representative features in distinctive categories, illustrated in Figure \ref{fig:fu}.

In the GFU, the upsampling operation is performed first in order to recover feature maps from the size $(\frac{k_{i}}{s} ,  \frac{k_{i}}{s}, F)$ to $(k_{i}, k_{i}, F)$. Since the total size of the recovered feature map is extensive, vacant positions in the matrix are filled with zeros as they do not affect the graph semantics in corresponding feature dimensions. Next, a transformation operation is performed in order to convert each 3D feature map back to a 2D sub-feature matrix $\boldsymbol{X'}$. Because nodes are assigned by the feature similarity in the GFP module, they are sorted in their original order in each sub-feature matrix. Meanwhile, a corresponding mask with the same matrix shape with a special value (0/1) is also generated to record the position of the local representative features. A value of 1 in each mask indicates a local representative attribute at this position in the corresponding sub-feature matrix. In contrast, a value of 0 represents the corresponding value at this position may be unrepresentative in this category.

Since the local representative attributes have already been determined in each updated feature matrix, they can now be utilized to find the internal priority of nodes by category. In this step, a learnable self-attention scheme is applied to each sub-feature matrix in Equation \ref{feature_self_attention}.

\begin{ceqn}
\begin{equation}\label{feature_self_attention}
\alpha_{j} = softmax(\mathbf{x_{j}}) =  \frac{exp(\mathbf{x_{j}})}{\sum_{j\subseteq f}exp(\mathbf{x_{j}})}
\end{equation}
\end{ceqn}

where $\mathbf{x_{j}}$ indicates the vector of $j$-th dimension in each sub-feature matrix.  

%下面这段有点啰嗦，尝试着精简下，顺便提出这是第三处可以降低计算量的方法
In Equation \ref{feature_self_attention}, current sub-feature matrices are still dense with different semantic values resulting in a heavy computation burden, even though most of these values are eventually replaced with zeros after upsampling. Since only the local dominant and representative features are needed, the computational burden can be reduced by using masks that record the position of local representative features, allowing the unnecessary zeros to be ignored in the computation.

%Although most values are replaced with zeros after the upsampling operation, they are still included in the current calculation, which results in a heavy computation burden. Since these zeros are unrepresentative in each category, there is no need to execute this calculation. Only those local dominant and representative features should be used to calculate the internal contribution of node attributes in sub-feature matrices under the node classification task. Since masks are recorded the position of local representative features, we can use them to further reduce the computational burden in this step. As shown in Figure \ref{fig:sparse}, each sub-feature matrix is converted from a dense matrix to a sparse one. The masks help reduce redundancies in the dense matrix for the computation reduction. 

\begin{figure}[t] 
\begin{center} 
\includegraphics[width=\linewidth]{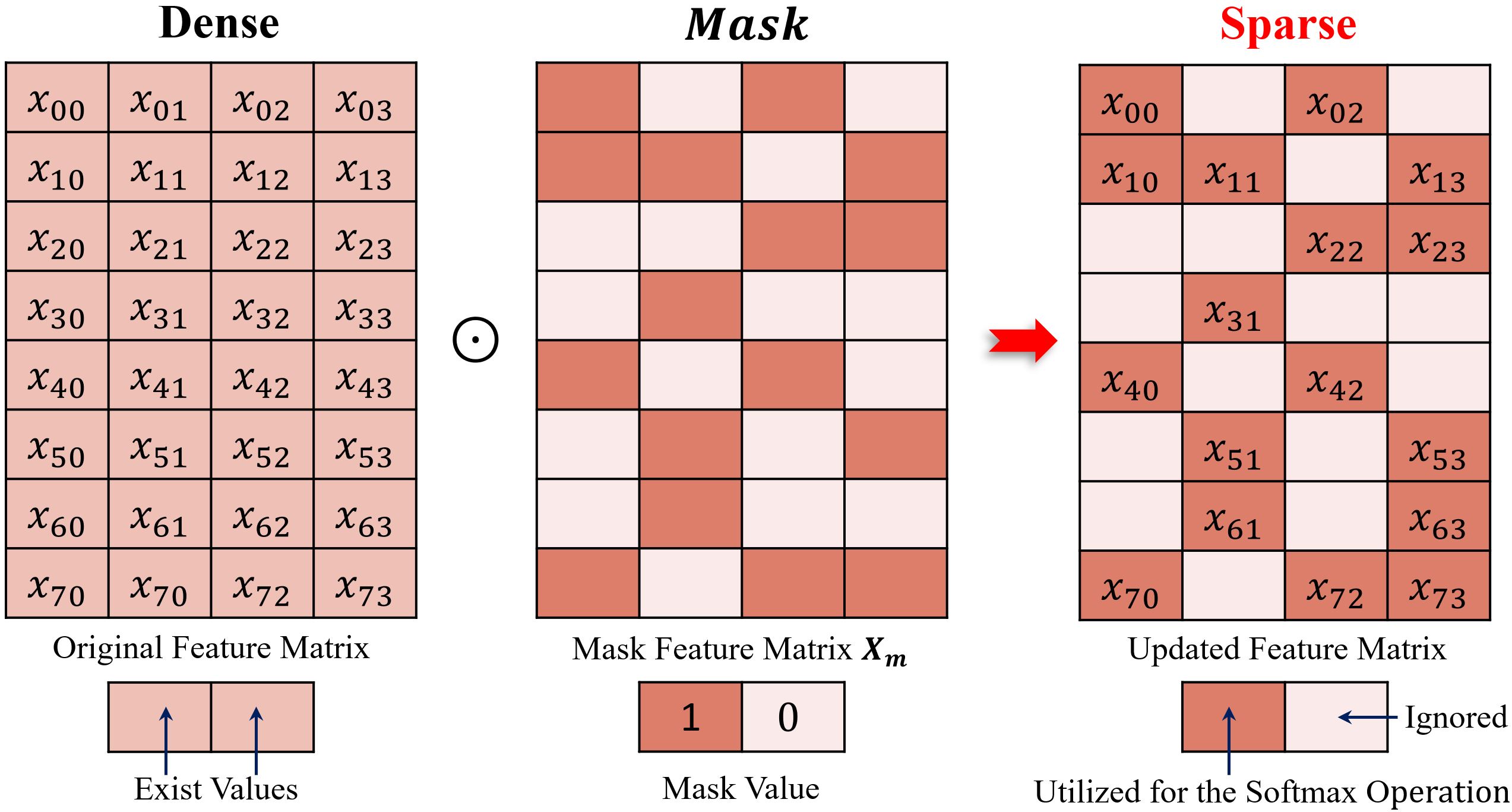} 
\end{center} 
\caption{Each dense sub-feature matrix is converted to a sparse matrix by applying a corresponding mask. A value of 1 in each mask indicates a local dominant and representative feature at this position in the corresponding sub-feature matrix. In order to reduce the computation burden, only these representative features in the sparse matrix are computed in future operations.} \label{fig:sparse}
\end{figure}

%https://www.overleaf.com/learn/latex/Algorithms
\begin{algorithm}\label{algo}
\SetAlgoLined
\small
\DontPrintSemicolon
\SetKwInOut{Input}{Input}
\SetKwInOut{Output}{Output}
 \Input{The original 2D feature matrix $\boldsymbol{X}$}
 \Output{The updated 2D feature matrix $\boldsymbol{X'}$}
 1. \textbf{Split} $\boldsymbol{X}$ to $n$ sub-feature matrices // $n = C + 1$\;
 \hspace*{3mm}// for each category\;
 2. \For{$i\gets1$ \KwTo $n-1$ \do{} }{
    \vspace*{1.5mm}\hspace*{3mm} // GFP module\; \vspace*{0.5mm}
    3. \textbf{Mean} feature vector:  $\overline{\mathbf{x}}_{i} = \frac{1}{m_{i}}\sum_1^{m_{i}}\mathbf{x}$\;
    4. \textbf{Similarity}: $sim(\overline{\mathbf{x}}_{i}, \mathbf{x}) = cos\frac{\overline{\mathbf{x}}_{i} \bigodot \mathbf{x}}{\left \|\overline{\mathbf{x}}_{i}  \right \| \cdot  {\left \|\mathbf{x}  \right \|}}$ \;
    5. \textbf{Sort} the nodes based on the sim\;
    6. \textbf{Transform} to 3D feature maps \;
    \hspace*{3mm} $(m_{i} \times F)  \mapsto (k_{i} \times k_{i} \times F)$
    where $k_{i}=\lceil\sqrt{m_{i}}\rceil$\;
    7. \textbf{Max pooling} $(k_{i} \times k_{i} \times F) \mapsto (\frac{k_{i}}{s} \times \frac{k_{i}}{s}  \times F)$\;
    \vspace*{1.5mm} \hspace*{3mm} // GFU module\; \vspace*{1mm} 
    8. \textbf{Up sampling} $(\frac{k_{i}}{s} \times \frac{k_{i}}{s}  \times F) \mapsto (k_{i} \times k_{i} \times F) $\;
    9. \textbf{Convert} back to 2D sub-feature matrix\;
    \hspace*{3.5mm} $ (k_{i} \times k_{i} \times F) \mapsto (m_{i} \times F)$\;
    10. \textbf{Mask} to indicate the representative features\;
    11. Feature\textbf{ Attention } Coefficient with mask\;
    \hspace*{4mm} $\alpha_{j} = softmax(\mathbf{x_{j}})$\, where $j \gets (1, F)$;
 }
 11. \textbf{Merge} all sub-feature matrices to the matrix $\boldsymbol{X'}$

\caption{The dimension-based self-attention mechanism on the feature attention branch}
\end{algorithm}

\begin{table*}[h]
\centering
\caption{Details of graph datasets in our experiments.}
\vspace{0.5em}
\footnotesize
\begin{tabular}{c|ccccccc}
\hline\noalign{\small}
Dataset & \#Classes & \#Features & \#Nodes & \#Edges & \#Train nodes & \#Val nodes & \#Test nodes\\
\noalign{\small}\hline\noalign{\small}
Cora & 7 & 1433 & 2485 & 5069 & 140 & 500 & 1000 \\
Citeseer & 6 & 3703 & 2110 & 3668 & 120 & 500 & 1000 \\
PubMed & 3 & 500 & 19717 & 44324 & 60 & 500 & 1000 \\
Coauthor CS & 15 & 6805 & 18333 & 81894 & 300 & 500 & 1000 \\
Amazon Photo & 8 & 745 & 7487 & 119043 & 160 & 500 & 1000 \\
Coauthor Physics & 5 & 8415 & 34493 & 247962 & 100 & 500 & 1000 \\
Amazon Computers & 10 & 767 & 13381 & 245778 & 200 & 500 & 1000 \\
\noalign{\small}\hline
\end{tabular}
\label{tab:dataset}
\end{table*}

The LeayReLU non-linearity is considered as the activation function. Thus, the Equation \ref{feature_self_attention} can be updated as:

\begin{ceqn}
\begin{equation}\label{feature_attention_update}
\alpha_{j}  =  \frac{exp(LeakyReLU(\theta, \mathbf{x_{j}}))}{\sum_{k\subseteq N_{i}}exp(LeakyReLU(\theta, \mathbf{x_{j}}))}
\end{equation}
\end{ceqn}

Where $\theta$ indicates the optimal parameters of the network.

The new non-linear node $v$'s feature is applied by the activation function $\varphi$ in Equation \ref{feature_attention}.

\begin{ceqn}
\begin{equation}\label{feature_attention}
FAB: \mathbf{x_{v}} = \varphi  \left ( \sum_{j\subseteq (1,F)} \alpha_{j} W \mathbf{x_{v}}\right )
\end{equation}
\end{ceqn}

Since all feature attention matrices by categories are adopted, the feature matrix $\boldsymbol{X'}$ is generated by merging these updated matrices with the last sub-feature matrix in their original order, as shown in Figure \ref{fig:fu}. The strategy of the feature attention mechanism is summarized in Algorithm \ref{algo}.

The final step is capturing both attention matrices from parallel branches on the graph. The interaction of ($\boldsymbol{NAB}$) and ($\boldsymbol{FAB}$) are combined to form an integrated attention matrix, ($\boldsymbol{IA}$), in Equation \ref{feature_merge}.

\begin{ceqn}
\begin{equation}\label{feature_merge}
\boldsymbol{h} =  \phi  \left [ W \cdot
\sigma \left ( \sum_{s\subseteq N_{v}} \beta_{vs} W \mathbf{x_{s}}\right ) \cdot \varphi  \left ( \sum_{j\subseteq (1,F)} \alpha_{j} W \mathbf{x_{v}}\right )\right ]
\end{equation}
\end{ceqn}

This combined attention matrix contains the attention information of the graph structure and the internal attributes of the nodes needed to complete a single head prediction.

\begin{table*}[h]
\centering
\footnotesize
\caption{Comparison of the performance (accuracy\%) of different algorithms for the node classification task on different datasets.}
\vspace{0.5em}
\begin{tabular}{c|ccccccc}
\hline\noalign{\small}
Method & Cora & Citeseer & PubMed & Coauthor CS & Coauthor Physics & Amazon Computers & Amazon Photo\\
\noalign{\small}\hline\noalign{\small}
MLP & 55.1 & 46.5 & 71.4 & 88.3 & 88.9 & 45.1 & 69.6 \\
LabelProp & 73.9 & 66.7 & 72.3 & 76.7 & 86.8 & 75.0 & 83.9 \\
MoNet & 81.7 & 71.2 & 78.6 & 90.8 & 92.5 & \textbf{83.5} & 91.2 \\
GCN & 81.5 & 70.3 & 79.0 & 91.1 & 92.8 & 82.6 &  91.2 \\
GraphSAGE & 79.2 & 71.6 & 77.4 & 91.3 & 93.0 & 82.4 & \textbf{91.4} \\
GAT & 83.4 & 72.5 & 79.0 & 90.5 & 92.5 & 78.0 & 85.1 \\
\textbf{GD} & \textbf{88.3} & \textbf{78.9} & \textbf{85.5} & \textbf{95.5} & \textbf{97.3} & 83.1 & 89.3\\
\noalign{\small}\hline
\end{tabular}
\vspace{-1em}
\label{tab:accuracy}
\end{table*}

\begin{figure*}[t]
\begin{center}
\includegraphics[width=\linewidth]{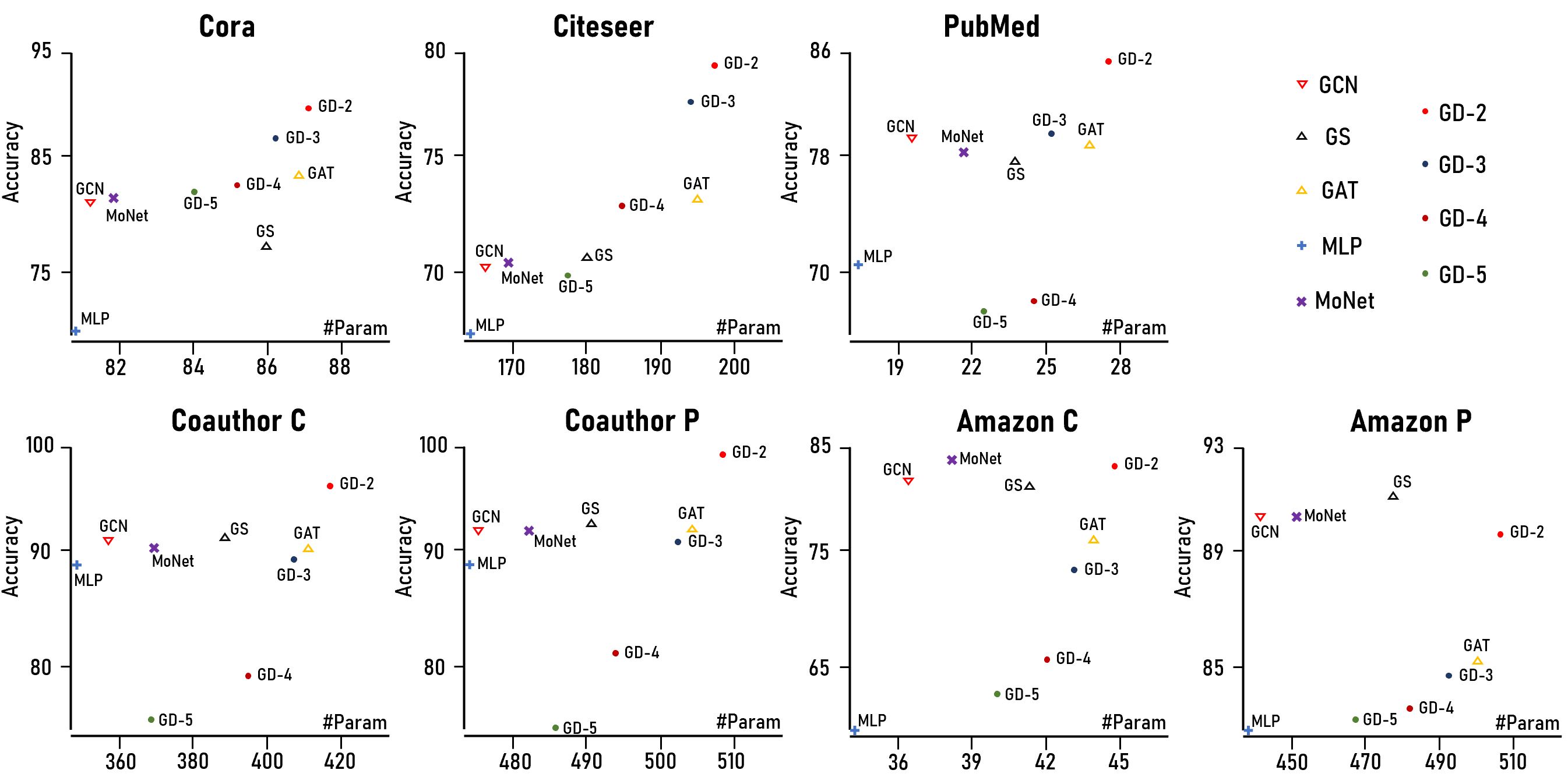}
\end{center}
\caption{The performance (accuracy\%) and parameters of different networks on seven datasets under node classification task. GS: GraphSAGE; GD-x is the Graph Decipher with the graph filter size of x.}
\label{fig:param}
\end{figure*}

\subsection{Global-Level: Multi-head Attention Mechanism}\label{mam}
The above sections describe the architecture of a single-head layer. The multi-head mechanism is finally used to stabilize the learning process under the node classification task, which is beneficial to GAT. Unlike GAT, the multi-head attention mechanism assigns distinctive attentions, $\mathbf{c}$, on each head in Equation \ref{multi-head}.

\begin{ceqn}
\begin{equation}\label{multi-head}
\boldsymbol{h_{l+1}}= \sigma(\frac{1}{\zeta}\sum_{\zeta=1}^{\zeta}\mathbf{c}\cdot \boldsymbol{h_{l}})
\end{equation}
\end{ceqn}

Where $\zeta$ is the number of multiple heads. Since the features are concatenated with distinctive head attentions, the average is calculated to make the final node classification with the loss function in cross-entropy.

\begin{table}[h]
\caption{This table displays the statistic of node attribute values in the above 7 graph datasets. The symbol, $57\natural$, indicates the PubMed dataset's 57 special fractions (feature values).}
\vspace{0.5em}
\centering
\footnotesize
\begin{tabular}{c|c}
\hline\noalign{\small}
Dataset & Feature Values\\\hline
\noalign{\small}\hline\noalign{\small}
Cora & [0, 1] \\
Citeseer & [0, 1] \\
PubMed & [   0, $57\natural$]  \\
Coauthor CS & [0 $\sim$ 5] \\
Amazon Photo & [0, 1] \\
Coauthor Physics & [0 $\sim$ 10, 14, 21, 28, 29, 37] \\
Amazon Computers & [0, 1] \\\hline
\noalign{\small}\hline
\end{tabular}
\label{tab:features}
\end{table}

\begin{table*}[h]
\centering
\footnotesize
\caption{The comparison of the performance (accuracy\%) of different algorithms for the node classification task on different datasets.}
\vspace{0.5em}
\begin{tabular}{c|ccccccc}
\hline\noalign{\small}
Method & Cora & Citeseer & PubMed & Coauthor CS & Coauthor Physics & Amazon Computers & Amazon Photo\\
\noalign{\small}\hline\noalign{\small}
MLP & 55.1 & 46.5 & 71.4 & 88.3 & 88.9 & 45.1 & 69.6 \\
LabelProp & 73.9 & 66.7 & 72.3 & 76.7 & 86.8 & 75.0 & 83.9 \\
MoNet & 81.7 & 71.2 & 78.6 & 90.8 & 92.5 & \textbf{83.5} & 91.2 \\
GCN & 81.5 & 70.3 & 79.0 & 91.1 & 92.8 & 82.6 &  91.2 \\
GraphSAGE & 79.2 & 71.6 & 77.4 & 91.3 & 93.0 & 82.4 & \textbf{91.4} \\
GAT & 83.4 & 72.5 & 79.0 & 90.5 & 92.5 & 78.0 & 85.1 \\
\textbf{GD} & \textbf{88.3} & \textbf{78.9} & \textbf{85.5} & \textbf{95.5} & \textbf{97.3} & 83.1 & 89.3\\
\noalign{\small}\hline
\end{tabular}
\vspace{-1em}
\label{tab:accuracy}
\end{table*}

\begin{figure*}[t]
\begin{center}
\includegraphics[width=\linewidth]{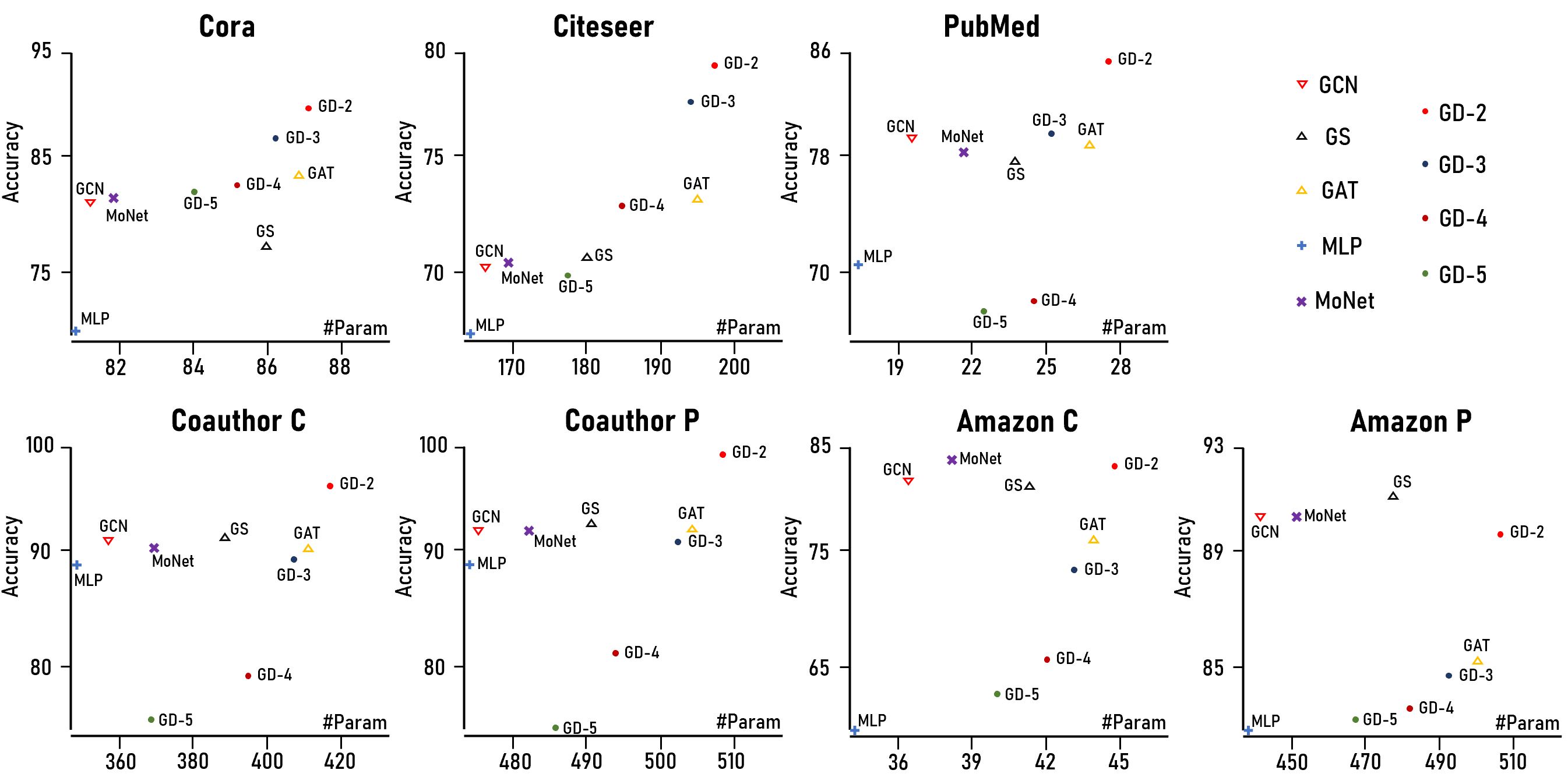}
\end{center}
\caption{The performance (accuracy\%) and parameters of different networks on 7 datasets under the node classification task. GS: GraphSAGE; GD-x shows Graph Decipher with a graph filter size of x.}
\label{fig:param}
\end{figure*}

\section{Experiments}\label{EX}

\subsection{Dataset}
We experimentally validate our proposed algorithm on seven real-world graph datasets, as shown in Table \ref{tab:dataset}, which summarizes the statistics of these seven datasets and the configuration of the train/val/test splits. The statistics summary of node attribute values in each dataset is shown in Table \ref{tab:features}.

\textbf{Cora and Citeseer} \cite{sen2008collective}, \textbf{PubMed} \cite{namata2012query}: These three public datasets are graphs used to describe citation patterns of scientific publications. The nodes represent publications, while the edges indicate the citation links among distinctive publications. For the Cora and Citeseer datasets, dictionaries (feature vectors) are utilized to explore the most common words that appear in these publications. Thus, each publication is described by a 0/1 value, which indicates the absence/existence of the corresponding word from the dictionary. While for the publications in the PubMed dataset, a term frequency-inverse document frequency (TF/IDF) is used to calculate the separation between them. In summary, the Cora dataset includes 2708 nodes with 5429 links in 7 categories, and the dimension of each node feature is 1433. The Citeseer dataset consists of 3327 nodes with 4732 links in 6 categories, and each node feature has 3703 dimensions. The PubMed dataset contains 19717 nodes with 44338 links in 3 categories and 500 dimensions per node feature vector.

\textbf{Amazon Computers and Amazon Photo} \cite{mcauley2015image}: These two datasets represent two different Amazon co-purchase graphs. Each node denotes products in different categories, while edges show two interests in bundle sales. And each dimension of the node features represents bag-of-words encoded product reviews. In summary, the Amazon Computers dataset includes 11381 nodes with 245778 links in 10 categories, and the dimension of each node feature is 767. On the other hand, the Amazon Photo dataset consists of 7487 nodes with 119043 links in 8 categories, and each node feature has 745 dimensions.

\textbf{Coauthor CS and Coauthor Physics} \cite{shchur2018pitfalls}: These two datasets are co-authorship graphs. Each node represents the authors, while edges indicate these two nodes co-authored a paper. The authors are grouped into different active fields or categories. Node feature vector illustrates the paper's keywords for each node, which represents the author's article. In summary, the Coauthor CS dataset includes 18333 nodes with 81894 links in 15 categories, and the dimension of each node feature is 6805. The Coauthor Physics dataset consists of 34493 nodes with 247962 links in 5 categories, and each node feature has 8415 dimensions.

%\begin{figure}[t]
%\begin{center}
%\includegraphics[width=\linewidth]{images/pool_size.jpg}
%\end{center}
%\vspace{-2em}
%\caption{The performances of GD with different graph feature pooling size on graph datasets.}
%\label{fig:pool_size}
%\end{figure}

%\begin{table*}[h]
%\centering
%\footnotesize
%\caption{The comparison of the performance (accuracy\%) of different graph feature pooling size in different dataset.}
%\vspace{0.5em}
%\begin{tabular}{c|ccccccc}
%\hline\noalign{\small}
%Size & Cora & Citeseer & PubMed & Coauthor C & Coauthor P & Amazon C & Amazon P\\
%\noalign{\small}\hline\noalign{\small}
%\textbf{2} & \textbf{88.3} & \textbf{78.9} & \textbf{85.5} & \textbf{95.5} & \textbf{97.3} & \textbf{83.1} & \textbf{89.3}\\
%3 & 86.4 & 77.5 & 81.1 & 88.9 & 91.1 & 73.7 & 84.4 \\
%4 & 82.9 & 72.6 & 66.1 & 79.3 & 80.9 & 67.3 & 79.0 \\
%5 & 81.8 & 69.5 & 65.8 & 73.0 & 74.2 & 61.1 & 75.9 \\
%\noalign{\small}\hline
%\end{tabular}
%\label{tab:gfp_size}
%\end{table*}

\subsection{Experimental Setup}
In order to avoid the gradient from exploding or vanishing during the learning process, we chose Xavier \cite{glorot2010understanding} to initialize the parameters of GD. The exponential linear unit (ELU) \cite{clevert2015fast} yields non-linear outputs at the end of both modules, NAB and FAB. The softmax \cite{ren2017robust} is used to send the probability distribution over predicted node categories at the end of GD. Moreover, during the training process, the dropout approach \cite{srivastava2014dropout} is introduced to avoid over-fitting, and the dropout rate is set to a range of $0.3$ to $0.7$ depending on the dataset. The graph feature pooling size, $s$, is set as 2, and the number of multi-heads, $\zeta$, is applied to 8.

% A circuit-breaker mechanism is employed to avoid validation accuracy vibrations over short ranges for extended periods. The best loss value is recorded and compared with the most recent value at the end of each epoch. If a lower value is recorded, this updates the best loss value. Simultaneously, if the epoch span for holding the latest record is larger than the predefined interval, such as 150 epochs, the circuit-breaker will terminate the training process. A circuit-breaker mechanism is used to prevent validation accuracy oscillating for extended periods with only negligible improvement. 

\subsection{Performance on Node Classification Task}

In our experiments, for each dataset, a fixed number of nodes from each class are selected for training and another 500 and 1000 nodes are utilized for validation and testing, as shown in Table \ref{tab:dataset}. The accuracy for the node classification task of distinctive algorithms on all seven datasets is illustrated in Table \ref{tab:accuracy}. It can be observed that the performance of the GNNs (MoNet \cite{monti2017geometric}, GCN, GraphSage, GAT, and ours) surpass the performance of non-GNN frameworks (MLP and LabelProp \cite{chapelle2009semi}), which benefits from the message-passing mechanism by considering both node attributes and graph structure on a graph.

The MoNet, GCN and GraphSage GNN's pass messages along edges uniformly on the graph. GAT evaluates only the contribution of direct neighbors to the central node on the graph structure. The measured performance of GAT is 83.4\%, 72.5\%, and 79.0\% on the Cora, Citeseer, and PubMed, respectively. It is superior to uniform GNNs on small graph datasets, but begins to degrade on large datasets with large numbers of node attributes, for example the Amazon Computers and Photos datasets. This performance decline is due to desired attributes being confined to edges along neighbor nodes which have a lower priority going to the center node, as all attributes are considered uniformly.
%This is because the desired attributes are restricted to be conveyed along edges from the neighbor nodes who has lower priority to the center node, since all node attributes are considered uniformly on this kind of GNN.

Because GD effectively biases its attention to representative node attributes alongside the most relevant neighbor nodes on the graph structure, it achieves state-of-the-art performance in 5 out of 7 datasets under the node classification task. GD achieves accuracies of 88.3\%, 78.9\%, 85.5\%, 95.5\%, 97.3\% on the Cora, Citeseer, PubMed, Coauthor CS, and Coauthor Physics datasets, respectively. In the case of the two largest datasets, Amazon Computer and Amazon Photo, GD lagged behind the frontrunner by only 0.4\% and 2.1\%, respectively. These experiments demonstrate the contribution of high priority components consisting of both node attributes and neighbors of the graph structure under the node classification task.

%The baselines of the latter consider only node attributes or graph structure. This demonstrates how the frameworks of GNNs are better suited for representation learning of graphs. Furthermore, it benefits from the message-passing mechanism of GNNs, which associates node attributes with graph structure in the information transmission procedure. 

%In Table \ref{tab:accuracy}, the graph feature pooling size is 2, and the number of multi-heads is chosen to be 8. Finally, in the ablation study sections, \ref{fs} and \ref{mh}, we present a comprehensive analysis of the above individual designs in the frame of our network.

\begin{table*}[h]
\centering
\footnotesize
\caption{Balance between complexity (MFLOPs) and accuracy of our network with distinctive multi-heads on Cora, Citeseer and PubMed datasets.}
\vspace{0.5em}
\begin{tabular}{c|r r||c|r r||c|r r}
    \hline
    \multirow{2}*{\textbf{Heads}} & \multicolumn{2}{c||}{\textbf{Cora}} & \multirow{2}*{\textbf{Heads}} & \multicolumn{2}{c||}{\textbf{Citeseer}} &
    \multirow{2}*{\textbf{Heads}} & \multicolumn{2}{c}{\textbf{PubMed}} \\
    & Flops(M) & AP &  & Flops(M) & AP &  & Flops(M) & AP \\ \hline
2	&	22.01	&	85.7	&	2	&	27.03	&	76.6	&	2	&	1166.64	&	83.3	\\
4	&	36.67	&	86.3	&	4	&	45.01	&	77.3	&	4	&	1945.18	&	83.9	\\
6	&	51.33	&	87.3	&	6	&	63.01	&	77.9	&	\textbf{6}	&	\textbf{2719.16}	&	\textbf{85.1}	\\
\textbf{8}	&	\textbf{66.01}	&	\textbf{88.3}	&	\textbf{8}	&	\textbf{81.01}	&	\textbf{78.9}	&	8	&	3498.94	&	85.5	\\
10	&	80.67	&	88.5	&	10	&	99.05	&	79.0	&	10	&	4277.67	&	85.7	\\
\hline\hline
\end{tabular}
\label{tab:heads3}
\end{table*}

\begin{table*}[h]
\centering
\footnotesize
\caption{Balance between complexity (MFLOPs) and accuracy of our network with distinctive multi-heads on Coauthor CS, Coauthor Physics, Amazon Computer, and Amazon Photo datasets.}
\vspace{0.5em}
\begin{tabular}{c|r r||c|r r||c|r r||c|r r}
    \hline
    \multirow{2}*{\textbf{Heads}} & \multicolumn{2}{c||}{\textbf{Coauthor C}} &
    \multirow{2}*{\textbf{Heads}} & \multicolumn{2}{c||}{\textbf{Coauthor P}} &
    \multirow{2}*{\textbf{Heads}} & \multicolumn{2}{c||}{\textbf{Amazon C}} &
    \multirow{2}*{\textbf{Heads}} & \multicolumn{2}{c}{\textbf{Amazon P}} \\
    & Flops(M) & AP  &  & Flops(M) & AP  &  & Flops(M) & AP  &  & Flops(M) & AP\\\hline
2	&	1008.62	&	93.7	&	2	&	3568.26	&	94.9	&	2	&	537.07	&	81.3	&	2	&	168.21	&	87.1	\\
4	&	1681.04	&	94.5	&	4	&	5982.16	&	95.8	&	4	&	895.12	&	81.9	&	4	&	281.36	&	87.9	\\
6	&	2353.46	&	95.0	&	6	&	8326.02	&	96.6	&	\textbf{6}	&	\textbf{1248.17}	&	\textbf{82.8}	&	\textbf{6}	&	\textbf{389.53}	&	\textbf{89.0}	\\
\textbf{8}	&	\textbf{3025.87}	&	\textbf{95.5}	&	\textbf{8}	&	\textbf{10709.76}	&	\textbf{97.3}	&	8	&	1611.41	&	83.1	&	8	&	526.70	&	89.2	\\
10	&	3698.29	&	95.5	&	10	&	13083.77	&	97.5	&	10	&	1969.26	&	83.2	&	10	&	616.88	&	89.2	\\
\hline\hline
\end{tabular}
\label{tab:heads4}
\end{table*}

\subsection{Ablation Study}\label{as}
In this section, a comprehensive analysis of our network is provided. Section \ref{fs} demonstrates the impact of graph feature pooling size for the network performance and computation, and section \ref{mh} discusses the effectiveness of multi-heads architecture based on the proposed single-head layer.

\subsubsection{Filter Size of the Graph Feature Pooling}\label{fs}

%3D矩阵的每一个channel代表着该类别下所有节点在当前特征维度下的对应值。在生成3D矩阵之前，模型已经在2D那儿完成了根据每一个类别下的的属性相关性排序。因此3D矩阵上相邻的节点所表现的feature相近。这一步，我们希望通过pooling的方法来粗略地找到节点在每一个特征维度下的比较重要的属性值。
In GFP, the graph feature filter's size is a significant parameter that affects GD's performance and computation burden as it determines the amount of local dominant and representative features by category in the learning process. The GD's performance as a function of filter size applied to our seven datasets is shown in Figure \ref{fig:param}. In these tests, the accuracy and number of parameters are indicators of a network's performance and computation burden. 

As shown in Figure \ref{fig:param}, the GD with feature pooling size 2 (GD-2) performs nearly as well as or better than most algorithms tested. The GD-2 algorithm typically imposes a slightly more significant computation burden than competing algorithms, though the gains in performance are clear. By computing only representative attributes in the graph feature pooling filter, the GD-2 algorithm can perform more efficiently. We found the optimal graph feature pooling size to be 2, effectively balancing the network's performance and computation burden. Filter size of 3 (GD-3) offers much lower computation burdens; however, this comes at the expense of the network performance, as shown in each test. These experiments show that the graph feature pooling module successfully preserves the representative attributes under the node classification task, achieving clear performance gains with greater network capacity.

\subsubsection{Multi-Heads Architecture}\label{mh}
% 集中讨论head的数量和accuracy和MFlops之间的关系。
This section demonstrates how the addition of heads impacts the overall performance of the network. We evaluated the performance and complexity of GD on all seven graph datasets as a function of heads under the node classification task, as shown in \ref{tab:heads3} and \ref{tab:heads4}. The floating-point operations per second (FLOPs) and average precision (AP) are network complexity and performance indicators. As heads were added, an improvement trend was observed to the point of diminishing returns, usually when around eight or ten heads are employed. On the Cora, Citeseer, PubMed, and Coauthor datasets, dual-head performance of GD is 85.7\%, 76.6\%, 83.3\%, 93.7\%, 94.9\% respectively, while the GAT achieves 83.4\%, 72.5\%, 79.0\%, 90.5\%, and 90.5\%. These experiments demonstrate GD's superior ability to push the upper limit of an existing network's performance. When using four and six parallel heads, the performance trend continues to improve dramatically. The 8-head configuration improves over the 6-head configuration by 1.0\% and costs 14.08M FLOPs on the Cora dataset. Beyond an 8-head configuration, the trend begins to subside, as there is only a marginal 0.2\% gain in performance at 10-heads while costing a substantial 14.66M FLOPs. These results indicate that the 8-head configuration of GD achieves high performance with an optimally balanced complexity on the Cora, Citeseer, and Coauthor datasets, and on the PubMed and Amazon datasets, the 6-head configuration of GD is the ideal option considering the trade-offs in performance and complexity.

\begin{table*}[h]
\centering
\caption{Distributions of node categories in different graph datasets are illustrated. The symbols $\S$ and $ \divideontimes $ represent the first 8 and following 7 categories of the Coauthor CS dataset. The third symbol $\star$ indicates the first 8 categories in the Amazon Computers dataset, containing another 2 categories with 2156 and 291 nodes separately.}
\vspace{0.5em}
\footnotesize
\begin{tabular}{c|cccccccc}
\hline\noalign{\small}
Dataset & \#1 & \#2 & \#3 & \#4 & \#5 & \#6 & \#7 & \#8\\\hline
Cora & 351 & 217 & 418 & 818 & 426 & 298 & 180 & ---\\
Citeseer & 249 & 590 & 668 & 701 & 596 & 508 & --- \\
PubMed & 4103 & 7739 & 7875 & --- & --- & --- & --- & ---\\
Coauthor CS $\S$ & 708 & 462 & 2050 & 429 & 1394 & 2193 & 371 & 924 \\
Coauthor CS $ \divideontimes $ & 775 & 118 & 1444 & 2033 & 420 & 4136 & 876 & ---\\
Amazon Photo & 369 & 1686 & 703 & 915 & 882 & 823 & 1941 & 331 \\
Coauthor Physics & 5750 & 5045 & 17426 & 2753 & 3519 & --- & --- & ---\\
Amazon Computers $\star$ & 436 & 2142 & 1414 & 542 & 5158 & 308 & 487 & 818 \\
\noalign{\small}\hline\noalign{\small}
\noalign{\small}\hline
\end{tabular}
\label{tab:imbl}
\end{table*}

\subsection{Graph Data Augmentation}
% 整个5.6需要重写！！

%1. 做data augmentation似乎是共识，那为什么别的network不行或者不好？你有什么优势？这个需要说明，而且要在5.6开头就着重强调和分析。

%2. 对于现有dataset的各类数据不平衡的情况要说明清楚。现在很混乱。

%3. 为什么只选取两个dataset，而且对两个dataset只选取两个类来实验？你肯定能找个理由回答，但我相信正常的reviewer都会要你补全实验。正常读者所期待的肯定是你找到了全新的强大的数据增强的方式，将原本不平衡的类都进行了平衡，然后再在所有的dataset上进行了验证。
In section \ref{rmi}, we discuss the significance of graph data augmentation and its influence on the multi-class imbalanced dataset. Without data augmentation, most false predictions are concentrated in minority categories in each graph dataset, which causes the network's performance to diminish. Current researches \cite{drummond2003c4, chawla2002smote, japkowicz2002class, shi2020multi} rely on attempts to balance the distribution of categories by sampling from a portion of the original datasets or class-conditioned adversarial graph learning. The primary issue with these approaches is that regardless of their relevance, all node features or attributes are considered in the graph learning process. It means insignificant node attributes are amplified and affect graph learning. We hope to evaluate only the critical node features or attributes to balance the node category distribution. Thus, we propose a new method of graph data augmentation to improve the network's performance, especially on minority categories, by utilizing the FAB of the message-passing mechanism.

\subsubsection{Distribution of Node Categories}\label{dnc}
This section summaries the distribution of the node categories of all seven datasets, as illustrated in Table \ref{tab:imbl}. 

\textbf{Cora}: 2708 nodes in 7 categories. Cora dataset holds the second least number of nodes among all seven datasets. The number of the top node class (\#4) is 818, while the smallest node class (\#7) only exists 180. The ratio between the majority and minority is around 9:2. 

\textbf{Citeseer}: 2110 nodes in 6 categories. The Citeseer dataset includes the least number of nodes among all seven graph datasets. The number of the top node class (\#4) is 701, while the smallest node class (\#1) only exists 249. The ratio of the majority and minority is around 3:1. 

\textbf{PubMed}: 19717 nodes in 3 categories. PubMed dataset contains the least types of the node category among all seven graph datasets. The number of the top node class (\#3) is 7875, while the smallest node class (\#7) only exists 4103. The ratio of the majority and minority is around 2:1.  

\textbf{Coauthor CS}: 18333 nodes in 15 categories. Coauthor CS dataset has the most types of the node category among all seven graph datasets. The number of the top node class (\#14) is 4136, while the smallest node class (\#10) only exists 118. The ratio of the majority and minority is around 35:1. Thus, the distribution of this graph dataset is the most imbalanced in our experiment.

\textbf{Coauthor Physics}: 34493 nodes in 5 categories. Coauthor Physics dataset obtains the most nodes among all seven graph datasets. The number of the top node class (\#3) is 17426, while the smallest node class (\#10) only exists 2753. The ratio of the majority and minority is around 6:1. 

\textbf{Amazon Photo}: 7487 nodes in 8 categories. The number of the top node class (\#7) is 1941, while the smallest node class (\#8) only exists 331. The ratio of the majority and minority is around 6:1.

\textbf{Amazon Computers}: 13381 nodes in 10 categories. Amazon Computers dataset involves the second largest number of node categories among all seven graph datasets. The number of the top node class (\#5) is 5158, while the smallest node class (\#10) only exists 291. The ratio of the majority and minority is around 18:1.  

\begin{figure*}[t]
\begin{center}
\includegraphics[width=\linewidth]{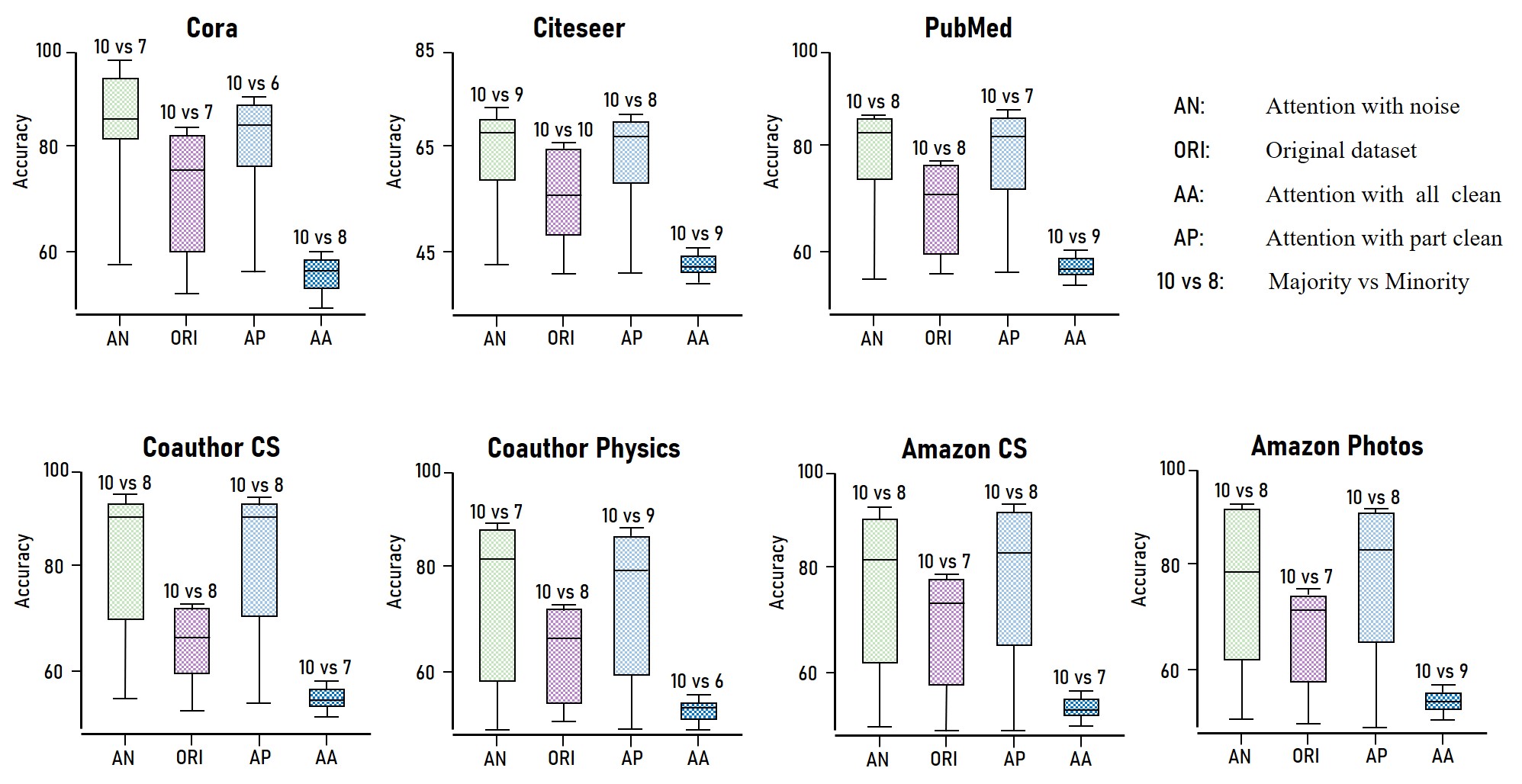}
\end{center}
\caption{The performance distributions of four data augmentation operations. The accuracy indicates the effect on performance of different approaches on the minority category. The numbers above the max bar, for example, 10 vs 7, represent the ideal proportion of the majority and minority categories on each inhomogeneous synthesized dataset. 
AN: Attention with Noise; Ori: Originally clone the information of minority.
AP: Attention with Part Clean;
AA: Attention with All Clean.}
\label{fig:mami}
\end{figure*}

\subsubsection{Data augmentation on two categories: majority and minority}\label{damm}
%先概况地说什么是不平衡问题
%Section \ref{dnc} describes the inhomogeneous distribution of the node category in all seven datasets. 
The multi-class imbalanced phenomenon generally causes most false predictions in the node category with the smaller data collection in the inference process. In order to prove our conjecture, nodes from the major and minor node categories are extracted to create a new imbalanced train and test datasets separately. The Cora and Citeseer datasets are used to test our retrained network on these new inhomogeneous datasets, which only contain two node categories. The network achieves 99.7\% and 98.9\% accuracy on these new imbalanced Cora and Citeseer datasets. This kind of `extraordinary' performance is because the node attributes with the major node category are weighted more heavily in the training process and propagated extensively on the imbalanced test dataset. However, if the retrained network only tests on the minority nodes, the accuracy drops to 47.6\% and 42.1\%, respectively, agreeing with our prediction. To solve this issue, we reconsidered the utilization of feature attention of the pre-trained model to gain efficiency. Therefore, we designed a series of experiments to find feasible data augmentation approaches to alleviate the imbalanced problem.

% 如何解决问题
In these experiments, each graph dataset is split into ten imbalanced datasets, containing only two node categories, majority and minority, with differing proportions. The proportion of majority and minority in the first imbalanced dataset is around 10:1. A straightforward approach for releasing the imbalanced node classification problem is to balance the distribution of nodes in different categories by reproducing the minority nodes and their corresponding attributes efficiently. The simplest method, ORI, is to clone the nodes and node attributes in the minority directly. However, it may lead to the induced over-fitting issue because the network amplifies and learns the inconsequential node features in the learning process. In order to solve this issue, another innovative approach for node reproduction are proposed.

%Before the reproduction, it is crucial and efficient to analyze the node attributes in all seven graph datasets. 
Since GD can explore the priority of node attributes under the node classification task, the node attributes are separated into two groups: representative features and unrepresentative ones. The former are retained in the reproduction, and the unrepresentative features are addressed with two different approaches to reproduce the minority: 1. AA: All the inconsequential attributes are cleaned, and then the cleaned node information is directly cloned. 2. AP: For each reproduction, some inconsequential attributes are cleared randomly. Thus, the reproduced data is not the same as the second approach. Finally, the propagation of majority and minority is followed by 10:1, 10:2, 10:3, 10:4, 10:5, 10:6, 10:7, 10:8, 10:9 and 10:10. The GD is then retrained on the newly synthesized datasets. 

In terms of the inference process, the test dataset now only includes the minority. The accuracy of the imbalanced synthesized datasets is summarized in Figure \ref{fig:mami}. The AA results arrangement is less than 60\% in all cases because all the inconsequential attributes are cleared. This result indicates that even inconsequential attributes have insignificant contributions to the node classification tasks. The maximum scores of the ORI approach increased in all cases. However, the scores of the ORI approach are only better than AA in one of our seven cases. While there are performance gains, improvements are still needed if the minority node information is repetitive. Furthermore, the AP approach's performance is extended compared to the ORI approach in every case because the distribution of the synthesized dataset with AP is diversified.

Following this discovery, another approach called AN is used to reproduce the dataset. In this approach, some small amounts of random noise is introduced to replace some inconsequential node attributes. An achieved higher scores than AP on the following datasets: Cora, Citeseer, Coauthor CS. However, in the remaining datasets, AP performed better than AN. The interquartile range (IQR) provides a visual indicator regarding the spread of accuracy amongst different synthesized datasets for each case. From Figure \ref{fig:mami}, the differences in IQRs for both AN and AP are tiny for each case, meaning both approaches can improve predictions on the minority category of the inhomogeneous dataset. However, it is essential to note that the artificial noise introduced does not have any practical use in real-world applications. Thus, although the network is more robust, the AN approach cannot be used in certain domains which require a strong interpretability graph network, such as in medical science. 

In addition, the number above the max bar in Figure \ref{fig:mami} represents the proportion of the majority and minority categories on each imbalanced synthesized dataset. Here the max scores of each approach are not from the most balanced setting, 10 vs 10, of the synthesized datasets in each case. For example, the best setting of AP in the Cora synthesized case is 10 (majority category) vs 6 (minority category). On the other hand, in synthesized Coauthor CS, the best proportion of the majority and minority categories is 10 vs 8 in both the AN and AP approach. This is because the synthesized datasets are generated based on a portion of the minority nodes. Thus, an appropriate number of generated nodes from the minority category is sufficient to balance the inhomogeneous multi-class issue.

\subsubsection{Data augmentation on all node categories}
\begin{figure}[t]
\begin{center}
\includegraphics[width=\linewidth]{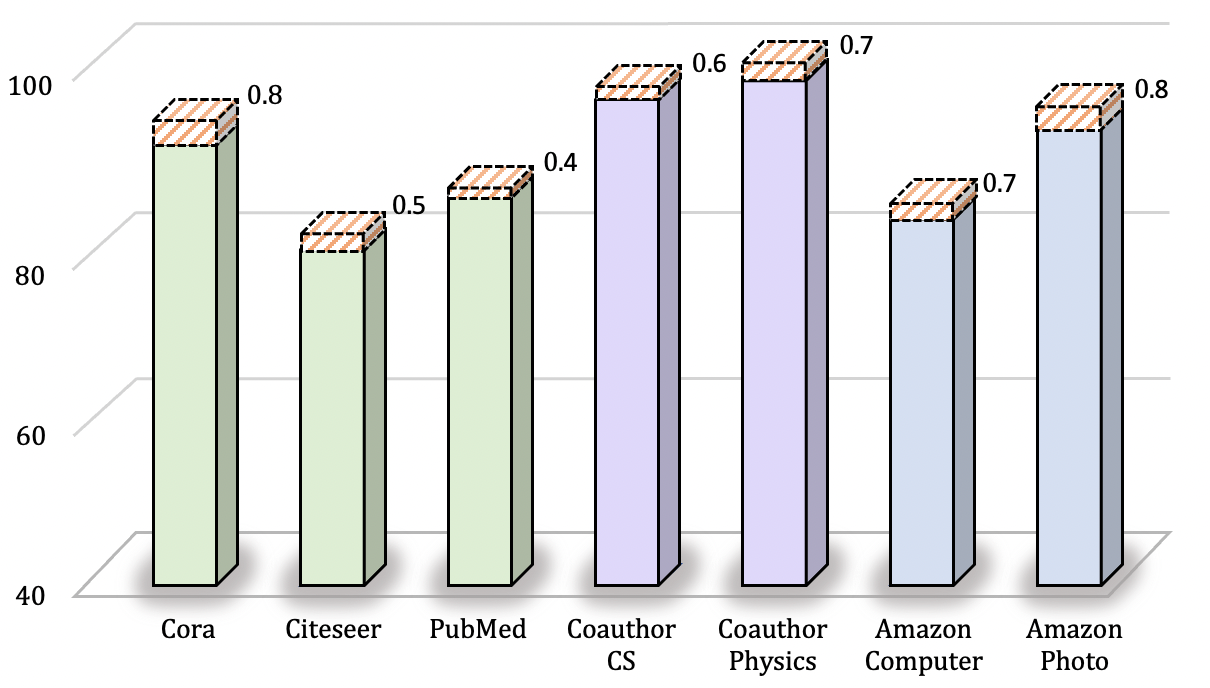}
\end{center}
\caption{Increment histogram. The y-axis indicates the accuracy of our network under the node classification task. The solid bar is the performance of our network on each original dataset. The shaded area shows the increment of network performance with the data augmentation approach, AP, on each graph dataset under the node classification task.}
\label{fig:ncda}
\end{figure}

%In section \ref{damm}, we demonstrated that our data augmentation approach could alleviate most false predictions that occur in the node category with the smaller data collection. However, since our approach has only been demonstrated on two categories in these datasets, 
This section is to demonstrate the performance of GD after data augmentation of all categories under the node classification task. We followed the same procedures in section \ref{damm} to balance the node distribution of all seven datasets: 1. Analyze all datasets by a pretrained GD; 2. Retain the representative features; 3. Reproduce the minorities by AP. After data augmentation, the ratio of each two categories is concentrated to 1:1.5 on each training dataset. Then, our network is retrained on new imbalanced graph datasets. \ref{fig:ncda} illustrates the increment of the network performance with the data augmentation. As a result, the accuracy of the network is further improved than the performance before augmentation on all datasets, which also demonstrates the improvements delivered by our innovative feature attention mechanism of GD.

\section{Conclusion}\label{CON}
In this paper, we propose a transparent GNN, Graph Decipher, that investigates the message-passing mechanism on a graph under the node classification task. GD improves functionality by showing how the graph structure and node attributes affect the message-passing mechanism in the node classification task from the graph, feature, and global levels. By giving higher priority to both neighbor nodes on the graph structure and representative features of node attributes, GD efficiently improves performance on the seven graph datasets studied. Meanwhile, the computation burden imposed by GD is acceptable due to three novel features: i. it explores the node attributes with category-oriented feature attention coefficients; ii. it investigates the representative attributes retained by the graph feature pooling filter; iii. it calculates the interior priority of node attributes on the sparse matrix generated from the mask. Additionally, an innovative GD-based graph data augmentation approach alleviates the imbalanced node classification problem on multi-class graph datasets. We hope that these discoveries will encourage future research into the possibilities of graph neural networks in additional real-world applications.

\textbf{Declaration of competing interest}
The authors declare that they have no known competing financial interests or personal relationships that could have appeared to influence the work reported in this paper.

\bibliography{mybibfile}

\end{document}